\documentclass[letterpaper, 10 pt, journal, twoside]{ieeetran}
\usepackage{cite}
\usepackage{amsmath,amssymb,amsfonts}
\usepackage{graphicx}
\usepackage{subcaption}
\usepackage{textcomp}
\usepackage[table,x11names]{xcolor}

\usepackage{enumitem}
\usepackage{optidef}
\usepackage[skip=2pt]{caption}
\usepackage{subcaption}
\usepackage{multirow}
\usepackage{algorithm}
\usepackage[noend]{algpseudocode}
\usepackage{hyperref}
\usepackage{soul}
\usepackage{placeins}
\usepackage[english]{babel}
\usepackage[autostyle, english = american]{csquotes}
\MakeOuterQuote{"}

\algrenewcommand\algorithmicrequire{\textbf{Input:}}
\algrenewcommand\algorithmicensure{\textbf{Output:}}

\newtheorem{proposition}{Proposition}
\newtheorem{definition}{Definition}[section]

\newenvironment{sketchproof}{%
  \par\noindent\textit{Proof Sketch: }\ignorespaces
}{%
  \hfill$\square$\par
}

\begin{document}

\title{Generating and Optimizing Topologically Distinct Guesses for Mobile Manipulator Path Planning with Path Constraints}

\author{Rufus Cheuk Yin Wong, Mayank Sewlia, Adrian Wiltz, Dimos V. Dimarogonas
\thanks{Manuscript received: April, 25, 2025; Revised August, 23, 2025; Accepted September, 27, 2025.}
\thanks{This paper was recommended for publication by Editor Aniket Bera upon evaluation of the Associate Editor and Reviewers' comments.}
\thanks{This work was supported by the Swedish Research Council, the Knut and Alice Wallenberg Foundation, and the Swedish Foundation for Strategic Research.}
\thanks{The authors are with the Division of Decision and Control Systems, School of EECS, Royal Institute of Technology (KTH), 100 44 Stockholm, Sweden {\tt\small [rcywong,sewlia,wiltz,dimos]@kth.se}}
\thanks{Digital Object Identifier (DOI): see top of this page.}
}
% \author{Author names omitted for anonymous review}
\markboth{IEEE Robotics and Automation Letters. Preprint Version. Accepted September, 2025}
{Wong \MakeLowercase{\textit{et al.}}: Generating and Optimizing Topological Distinct Guesses} 

\maketitle

\begin{abstract}
Optimal path planning is prone to convergence to local, rather than global, optima. This is often the case for mobile manipulators due to nonconvexities induced by obstacles, robot kinematics and constraints.
This paper focuses on planning under end effector path constraints and attempts to circumvent the issue of converging to a local optimum.
We propose a pipeline that first discovers multiple homotopically distinct paths, and then optimizes them to obtain multiple distinct local optima.
The best out of these distinct local optima is likely to be close to the global optimum.
We demonstrate the effectiveness of our pipeline in the optimal path planning of mobile manipulators in the presence of path and obstacle constraints.
\end{abstract}

\section{Introduction}
% Optimal path planning for robots involves finding the path for a robot which minimizes some cost function such as time or energy, and subject to the constraints of the robot and task at hand.
% Such optimal paths are desirable since they lead to more efficient and often, more natural and predictable motions.
% In the context of mobile manipulators,
% the complex kinematics, high degrees of freedom (DoF) and obstacle avoidance requirements make the optimal path planning problem even more challenging.
\IEEEPARstart{O}{ptimal} path planning for mobile manipulators is commonly done by formulating and solving a nonlinear program (NLP) using gradient-based optimization approaches.
One major challenge with this approach is that often the constraints introduced to the planning problem, such as obstacle avoidance, end effector path constraints, cause the NLP to be highly nonconvex.
This causes gradient based optimization approaches to only solve them to local optimality.
While solving nonconvex NLPs to global optimality in general is NP-hard, one potential mitigation is to generate multiple distinct local optima and choose the best among them.
This increases the likelihood of actually finding the global optimum.
In the sequel, we denote the optimum among multiple distinct local optima a \textit{multi-local optimum}.
% \begin{definition}[Multi-Locally Optimal]
% We call a solution to an optimization problem \textit{multi-locally optimal}
% if it is the optimal solution among multiple distinct local optima.
% \end{definition}

The challenge of obtaining a multi-locally optimal path is computing multiple distinct local optima since most research has only been conducted on finding a single local optimum \cite{zucker2013chomp},\cite{schulman2014motion}.
% Naively using random initial guesses, or random perturbations of a single initial guess would likely produce the same local optimum or lead to infeasibility.
Using the observation that the local optimum returned by gradient-based optimization approaches usually stays within the same homotopy class as the provided initial guess \cite{Lav06},
we propose a pipeline that first discovers homotopically distinct paths and then uses them as initial guesses for an NLP.
This allows for generating multiple distinct local optima, and subsequently finding the \textit{multi-local optimum}.

\begin{figure}
\captionsetup{skip=0pt}
  \begin{center}
    \includegraphics[width=\columnwidth]{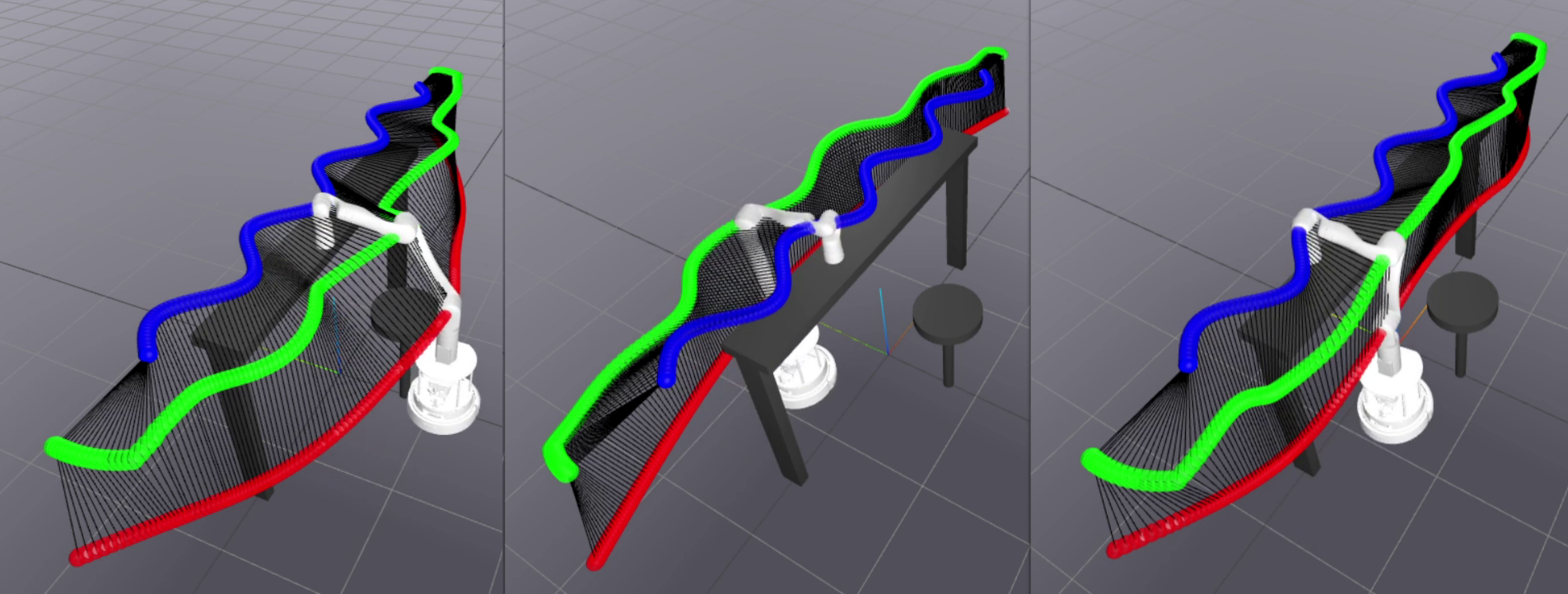}
  \end{center}
  \caption{Mobile manipulator executing three homotopically distinct locally optimal paths given a desired end effector path. Blue shows the desired end effector path, green and red shows the computed elbow and base paths respectively.}
  \label{fig:intro}
  % \vspace{-0.2cm}
\end{figure}

We apply our pipeline to the path planning of mobile manipulators consisting of a 6-revolute (6R) elbow manipulator attached to a nonholonomic differential drive base.
% This form was chosen for its cost-effectiveness and versatility.
% Such robots are cost-effective and versatile.
% The cost effectiveness and versatility of such mobile manipulators make them prime candidates for revolutionizing the way we work and live.
We further require that the end effector follows a predetermined path.
Such end-effector path constraints arise naturally in applications such as painting, welding or wiping a table.

The contribution of this paper is the development of a path planning pipeline for mobile manipulators under end effector path constraints and produces a \emph{multi-locally optimal} solution.
% This is achieved through the application of topological path planning and optimal path planning together.
To this end, we propose a method for generating a low dimensional configuration graph to be used with the Neighborhood Augmented Graph Search (NAGS) algorithm \cite{sahin2023topogeometrically}.
Additionally, we propose several modifications to the NAGS algorithm that enhances its accuracy.
Furthermore, an NLP is formulated to produce distinct locally optimal paths from the guesses provided by the modified NAGS algorithm.
Finally, the effectiveness of the pipeline is demonstrated with simulation results along with a comparison study with existing methodologies.

% To this end, our proposed pipeline involves leveraging the end effector path constraint to effectively generate distinct initial guesses for the mobile manipulator and finally optimizing over each guess to find the multi-locally optimal path.

% The planning pipeline consists of four main steps, illustrated in Fig. \ref{fig:pipeline}.

% \begin{figure*}[tbp]
%   \begin{center}
%     \includegraphics[width=\linewidth]{figures/pipeline.png}
%   \end{center}
%   \caption{The planning pipeline}
%   \label{fig:pipeline}
% \end{figure*}

% \begin{enumerate}
%   \item The collision-free configuration space graph is generated.
%   This is a graph where each node represents an obstacle-collision-free robot configuration
%   and each edge represents a collision-free transition between robot configurations.
%   This is created by discretizing the configuration space and testing each node and edge for obstacle collision.
%   \item The NAGS algorithm proposed in \cite{sahin2023topogeometrically} takes as input the collision-free configuration space graph
%   and finds topologically distinct paths within the graph.
%   \item A NLP is formulated encoding the nonholonomic base constraint, obstacle avoidance constraint and end effector path constraint,
%   as well as an objective to minimize the total path length.
%   The topologically distinct paths from the previous step are passed to an NLP solver as initial guesses and the NLP is solved.
%   \item The optimized paths from the different initial guesses are compared, and the best path is returned.
% \end{enumerate}

The remainder is organized as follows.
In Section \ref{sec:related_work}, related work is reviewed, and in Section \ref{sec:problem}, the problem under consideration is stated.
Section \ref{sec:methodology} presents the proposed planning pipeline in detail, and Section \ref{sec:results} presents some experimental results demonstrating the efficacy of our pipeline.
Section \ref{sec:conclusion} offers a conclusion with some discussion.

\section{Related Work}
\label{sec:related_work}
% \subsection{Mobile Manipulators}
\subsection{Constrained Motion Planning}
% Path planning for mobile manipulators has been widely studied.
% The survey \cite{sandakalum2022motion} provides a comprehensive overview on mobile manipulators planning algorithms.
% We highlight a few results in the context of planning with end effector path constraints on nonholonomic mobile manipulators.

% A stochastic rapidly exploring random tree (RRT) \cite{lavalle1998rapidly} based approach is introduced in \cite{oriolo2005motion} that handles constraints via tangent space projection.
% This has been further generalized and incorporated into OMPL \cite{kingston2019exploring}.
% % Such sampling approaches can often be postprocessed via a trajectory optimization framework to produce a locally optimal path.
% A stochastic genetic algorithm is proposed in \cite{vannoy2008real} which eventually produces an optimal solution.
% An inverse kinematics (IK) based method is proposed in \cite{pin1997including} which produces an optimal solution greedily.
% While all of these approaches can produce locally optimal paths directly or with postprocessing, by the nature of these algorithms (stochastic or greedy), there is no guaranteed way of discovering more than one unique local optimum.

Motion planning in high dimensional space such as mobile manipulators under end effector constraints has been well studied.
Many current methods build upon the rapidly exploring random tree (RRT) \cite{lavalle1998rapidly} and encode the constraints geometrically during tree construction \cite{oriolo2005motion,berenson2011constrained,kim2016tangent,jaillet2012path}.
% Similar constrained sampling based path planners include CBIRRT2 \cite{berenson2011constrained}, TB-RRT \cite{kim2016tangent}, AtlasRRT \cite{jaillet2012path}.
These have been generalized and incorporated into the Implicit Manifold Configuration Space (IMACS) framework \cite{kingston2019exploring}, decoupling the planning algorithm from the constraint adherence.
These sampling approaches can often be postprocessed to produce a locally optimal path.
% A genetic algorithm is proposed in \cite{vannoy2008real} which is probabilistically optimal but may take excessive amounts of time to achieve global optimality. An inverse kinematics (IK) based method is proposed in \cite{pin1997including} which produces a locally optimal solution greedily. None of these approaches provide a guaranteed way of discovering more than one distinct local optimum.
However, the randomized nature of RRT-based algorithms implies that there is relatively little control over the characteristics of the paths returned, e.g. the topological properties.

\subsection{Optimal Path Planning}
Trajectory optimization is a commonly used technique in optimal path planning.
This involves formulating the path-finding problem as a mathematical program
with costs and constraints, which is then solved with an optimizer.
This field is well studied with many successful algorithms
such as CHOMP \cite{zucker2013chomp} and TrajOpt \cite{schulman2014motion}.
Both of these approaches use a direct transcription based technique \cite{underactuated}, which involves discretizing the trajectory into a fixed number of discrete samples.
% The mathematical program is then expressed as
% \begin{mini*}|s|
%   {\substack{x[\cdot], u[\cdot]}} {
%     l_f(x[N]) + \sum_{n=0}^{N-1} l(x[n], u[n])
%   }{}{}
%   \addConstraint{x[n+1]}{=f(x[n], u[n])}
%   \addConstraint{x[0]}{=x_0}
%   \addConstraint{x[N]}{=x_g}
%   \addConstraint{}{+\text{ additional constraints}}
% \end{mini*}
% where $l_f(x)$ is the final cost on the state and $l(x, u)$ is a per step cost on the state and input.
These approaches generally scale well with the number of decision variables and constraints.
However, the presence of nonconvex constraints and cost functions leads to results that are locally, not globally optimal, and heavily dependent on the initial guess provided.

\subsection{Topological Path Planning}
\begin{figure}[tbp]
  \begin{subfigure}[b]{0.4\linewidth}
    \centering
    \includegraphics[width=0.95\columnwidth]{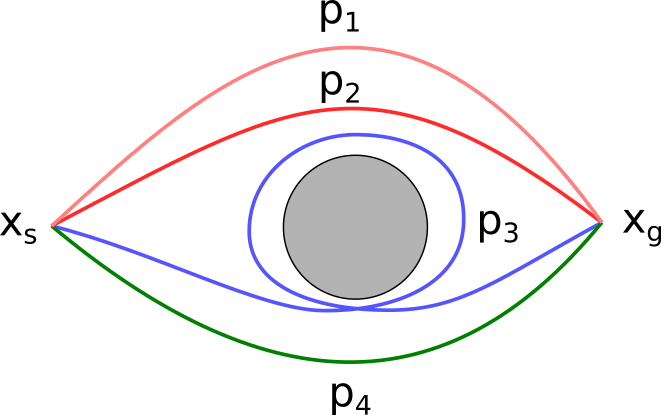}
    \caption{}
    \label{fig:homotopies}
  \end{subfigure}%
  ~
  \centering
  \begin{subfigure}[b]{0.55\linewidth}
    \centering
    \includegraphics[width=\textwidth]{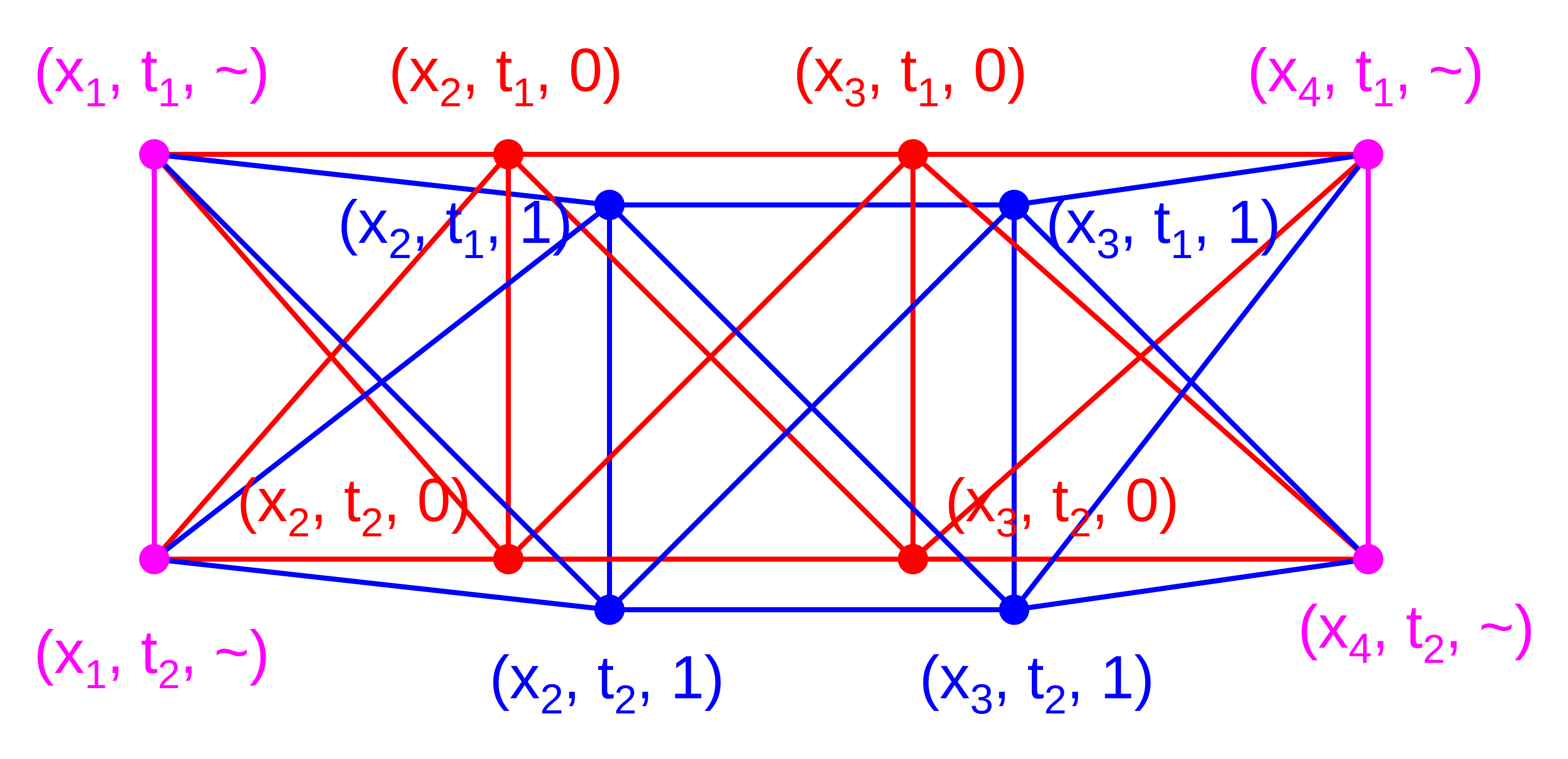}
    \caption{}
  \label{fig:1d_configuration_graph}
  \end{subfigure}%
  
  % \begin{subfigure}[b]{0.6\linewidth}
  %   \centering
  %   \includegraphics[width=\textwidth]{figures/1d_configuration_graph_combined.png}
  %   \caption{}
  %   \label{fig:1d_configuration_graph_combined}
  % \end{subfigure}  
  % \vspace{-0.2cm}
  \caption{(a) Given the grey obstacle, $p_1$ and $p_2$ belong to the same $\mathcal{H}$-class (homotopically equivalent) while $p_2$, $p_3$, $p_4$ each belong to a different $\mathcal{H}$-class (homotopically distinct).
  (b) Illustration of a 2D cross section ($y$ axis omitted) of the configuration graph.
  % Red and blue vertices are vertices within elbow configuration $w = 0$ and $w=1$, respectively. Purple vertices are vertices at the joint singularity.
  % This illustrates the effect of ``gluing'' the vertices at the boundaries together, as a consequence of changing elbow configuration at the joint singularity.
  }
\end{figure}

Topological path planning focuses on finding and quantifying paths based on their topological features.
Often, the feature of interest is a path's homotopy class ($\mathcal{H}$-class) within a robot's configuration space.
Paths of different $\mathcal{H}$-class cannot be smoothly deformed into each other without colliding with obstacles (Fig. \ref{fig:homotopies}).
Many probabilistic methods for finding homotopically distinct paths have been proposed \cite{pokorny2016high,bhattacharya2012topological,jaillet2008path}.
However, they generally scale poorly to high dimensions.
% Configuration space describes the space of feasible robot configurations.
% In the context of mobile manipulators, this is often described by the position of the mobile base and the pose of the robot arm.
% The homotopy class of paths in configuration space is then defined as follows:
% \begin{definition}[Homotopy Classes of Paths \cite{sahin2023topogeometrically}]
% Two paths connecting the same start and goal points in a configuration space are said to be in the same homotopy class ($\mathcal{H}$-class) or homotopic, if one can be continuously deformed into another
% without intersecting/crossing obstacles. Otherwise they are called nonhomotopic or homotopically distinct.
% \end{definition}

% This is illustrated in Fig. \ref{fig:homotopies}.
% \begin{figure}[tbp]
%   \begin{center}
%     \includegraphics[width=0.4\columnwidth]{figures/homotopies.png}
%   \end{center}
%   \caption{Given the grey obstacle, $p_1$ and $p_2$ belong to the same $\mathcal{H}$-class (homotopically equivalent) while $p_2$, $p_3$, $p_4$ all belong to different $\mathcal{H}$-classes (homotopically distinct).}
%   \label{fig:homotopies}
%   % \vspace{-0.2cm}
% \end{figure}

As such, using a lower dimensional or simpler topological path planning setup for coarse global planning followed by optimal path planning approaches for local refinement is a common approach to combine the best of both worlds.
This pipeline is effective in generating optimal trajectories for mobile ground robots \cite{rosmann2017integrated,he2022homotopy}, quadrotors \cite{zhou2021raptor} and manipulators \cite{rice2020multi,saleem2021search}.
% Particularly for manipulators, \cite{rice2020multi} provides an IK-based solution for finding optimal paths.
% However, it assumes a manipulator with one degree of redundancy, only considers distinct topologies caused by kinematic singularities and requires some way of identifying bifurcation points in the self-motion manifold.
% \cite{saleem2021search} utilizes a graph search technique combined with a topology heuristic for finding the optimal path.
% However, it requires projecting the 3D environment onto a 2D plane.

For applying this topological and optimal path planning pipeline to mobile manipulators, a major problem is the determination of the $\mathcal{H}$-class, which is very challenging for high-dimensional configuration spaces \cite{sahin2023topogeometrically}.
A novel Neighborhood Augmented Graph Search (NAGS) algorithm \cite{sahin2023topogeometrically} has recently been proposed that allows finding topologically distinct paths in higher dimensions.
% While the algorithm allows discovering homotopically distinct paths for up to 3 or even 4 dimensions,
% it still scales poorly with higher dimensions.
% As such, a combination of NAGS and optimal path planning suggests a way to achieve the best of both worlds.
In our work, we leverage a modified version of NAGS to identify homotopically distinct paths and use optimal path planning for the local refinement.

% This is the approach adopted in this work.

% To the best of our knowledge, topological path planning to optimal path planning pipeline has not been applied to mobile manipulators path planning yet.

\section{Problem Formulation}
\label{sec:problem}

The path planning problem concerns a 6-degree-of-freedom (DoF) fully actuated elbow manipulator attached to a nonholonomic differential drive mobile base.
The manipulator is assumed to have a 3DoF spherical wrist that handle any end effector orientation constraints.
This is the case for most mobile manipulators available today.
As such, we refer to the wrist as the end effector and only consider its position.

The base is characterized by its position $x_b = [x, y]^T \in \mathbb{R}^2$ and orientation $\theta \in S^1$.
The base motion is governed by
\begin{equation}
\dot x_b = \begin{bmatrix}\cos \theta \\ \sin \theta \end{bmatrix} u_1, \quad \dot \theta = u_2
\label{eq:base_dynamics}
\end{equation}
where $u_1, u_2 \in \mathbb{R}$ are control inputs.

As opposed to the more common approach of describing the arm in joint angle coordinates,
we instead choose to express the arm in maximal coordinates \cite{underactuated}.
In maximal coordinates, links are described by their position in space.
This allows for a more natural incorporation of the end effector constraint that we will exploit in Section \ref{sec:graph}.
Define $x_{b\perp} = [x_b, 0]^T = [x, y, 0]^T$.
Given the base position $x_b$, the arm can be characterized by its elbow position $x_w \in \mathbb{R}^3$ and end effector position $x_e \in \mathbb{R}^3$, both in world cartesian coordinates.
Let $l_1$ be the upperarm length and $l_2$ be the forearm length.
The elbow and end effector positions are subject to the following kinematic constraints:
\begin{align}\begin{split}
    \lVert x_w - x_{b\perp} \rVert _2 &= l_1 \\
    \lVert x_w - x_e \rVert _2 &= l_2 \\
    \exists a, b \in \mathbb{R} : x_w - x_{b\perp} &= a (x_e - x_{b\perp}) + \begin{bmatrix}0 & 0 & b\end{bmatrix}^T
    \label{eq:kinematics}
\end{split}\end{align}
The last constraint states that base, elbow and end effector positions projected to the $xy$-plane are collinear.
This reflects the fact that the upperarm and elbow cannot roll.
The dynamics of the arm is given by
\begin{equation}
\dot x_w = u_3, \quad \dot x_e = u_4
\label{eq:arm_dynamics}
\end{equation}
where $u_3, u_4 \in \mathbb{R}^3$ are control inputs, subject to the constraints in Eq. (\ref{eq:kinematics}).
The robot configuration $q$ is then fully defined by
$$
q = \begin{bmatrix}x_b^T & \theta & x_w^T & x_e^T \end{bmatrix}^T
$$
subject to the aforementioned constraints.
Note that this approach of using maximal coordinates to encode end effector constraints is not specific to the 6-DoF elbow manipulator and can be applied to arms with different kinematics by adjusting Eq. (\ref{eq:kinematics}) and (\ref{eq:arm_dynamics}), correspondingly \cite{brudigam2024variational}.

% The direct kinematics of the end effector can be expressed by $f: \mathbb{R}^2 \times [0, 2\pi] \times \mathbb{R}^3 \to \mathbb{R}^3$
% $$
% x_e = f(p) = f\begin{pmatrix}x_b \\ \theta \\ q \end{pmatrix}
% $$
% where $x_e \in \mathbb{R}^3$ is the end effector world coordinates, $x_b = (x, y)^T \in \mathbb{R}^2$ is the world coordinates of the mobile base, $\theta \in [0, 2\pi]$ is the base heading, and $q \in \mathbb{R}^3$ is the joint values (base pan, base lift, elbow flex) of the robot arm up until the wrist.
% The dynamics are given by
% \begin{equation*}
%     \dot x_b = \begin{bmatrix}\cos \theta \\ \sin \theta \end{bmatrix} v, \quad
%     \dot \theta = \omega, \quad
%     \dot q = u
% \end{equation*}
% where $v \in \mathbb{R}$ is the base linear velocity, $\omega$ is the base angular velocity, and $u \in \mathbb{R}^3$ is the joint velocities.

Obstacles are assumed to be defined via an obstacle function $\text{obs}(q)$ which returns \textit{True} if and only if the given robot configuration $q$ is colliding with an obstacle.
% Such obstacle function can be defined, e.g. via the signed distance function as follows:
% $$
% \text{obs}(p) = \text{False} \iff \text{sd}(\mathcal{L}_i(p), \mathcal{O}_j) \ge 0 \quad \forall i, j
% $$
% where $sd(\mathcal{L}, \mathcal{O}$ is the signed distance function, computable via the enhanced GJK algorithm \cite{cameron1997enhancing}, $\mathcal{L}_i(p) \subset \mathbb{R}^3$ is the set of points occupied by robot link $i$ while in pose $p$ and $\mathcal{O}_j \subset \mathbb{R}^3$ is the set of points occupied by obstacle $j$.

% A desired end effector path in world coordinates $\hat x_e(t)$ is given with $t \in [0, 1]$ being the path parameter.
% Similar to \cite{oriolo2005motion}, the planning problem is to find a path $p(t) = [x_b(t), q(t)]^T$ such that
% \begin{enumerate}
%     \item $\hat x_e(t) = x_e(t) = f(p(t)), \forall t \in [0, 1]$
%     \item $\text{obs}(p(t)) = False, \forall t \in [0, 1]$
%     \item $p(t)$ minimizes $\int_0^1 dp(t)$
%     \item $p(t)$ is feasible w.r.t. the dynamics.
% \end{enumerate}

A desired end effector path is given in the form of a function $x_e(k)$ with $k \in [0, 1]$ being the normalized path parameter.
The planning problem, then is to find a feasible path, satisfying the kinematic and end effector path constraints, and which does not collide with obstacles, minimizing
\begin{equation}
\int_0^1 \lVert \mathbf{u}(t) \rVert_2 dt
\label{eq:cost}
\end{equation}
% The final solution will be of the form $Q = \{q(t_0), q(t_1)$, $\dots, q(t_T)\}$, $t_0 = 0$, $t_T = 1$ with $t_i$ equally spaced and $T$ being the number of path samples.

% Our solution will be of the form
% $$
% \{p(t_0), p(t_1), \dots, p(t_T)\}, t_0 = 0, t_T = 1
% $$
% with $t_i$ equally spaced.
% $T$ is a parameter for the number of path samples.
% The final continuous path can be retrieved by simple linear interpolation.
\section{Methodology}
\label{sec:methodology}
% \fxnote{Move definition of multilocally optimal here}

\begin{figure}[tbp]
\captionsetup{skip=-2pt}
  \begin{center}
    \includegraphics[width=\linewidth]{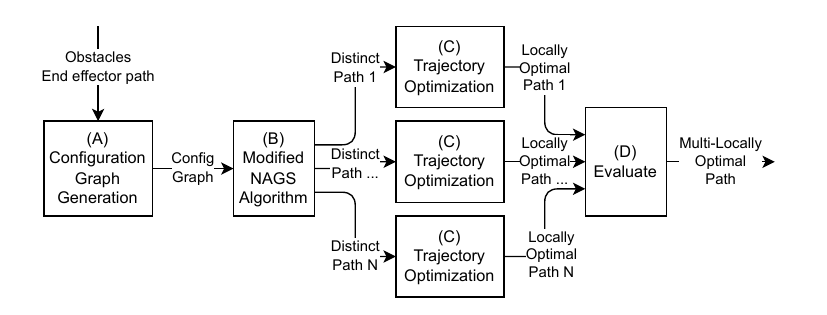}
  \end{center}
  \caption{The planning pipeline}
  \label{fig:pipeline}
\end{figure}

The proposed planning pipeline consists of four main steps, illustrated in Fig. \ref{fig:pipeline} and Algorithm \ref{alg:pipeline}.

\begin{enumerate}[label=(\Alph*)]
  \item First (line 2), we generate the collision-free configuration space graph (CG) representing the valid robot configurations and transitions.
  % Each vertex of the graph represents a collision-free robot configuration, and each edge represents a collision-free transition between configurations.
  \item Then (line 3), we apply a modified NAGS algorithm, adapted from \cite{sahin2023topogeometrically}, which takes as input the CG and finds a pre-specified number of homotopically distinct paths within the graph.
  \item Next (line 4-5), the homotopically distinct paths are used as initial guesses and the trajectory optimization problem is solved for each guess.
  \item Finally (line 6-7), we compare the optimized paths from the different initial guesses and choose the best path.
\end{enumerate}

\begin{algorithm}[t]
\caption{Pipeline for finding multi-locally optimal paths}
\label{alg:pipeline}
\begin{algorithmic}[1]
\Require
\Statex $x_e$: $[0, 1] \to \mathbb{R}^3$: Desired end effector path
\Statex $n$: Number of distinct local optima to evaluate
\Statex $dt$: Optimization timestep interval
\Statex $T$: Number of optimization timesteps
\Ensure $Q^\star = \{q(t_0), q(t_1)$, $\dots, q(t_T)\}$, $t_0 = 0$, $t_T = 1$
\Function{findPath}{$x_e, n, dt, T$}
\State $G = (V, E) := \text{ConfigurationGraphGeneration($x_e$)} $
\State $\begin{bmatrix}
    (x_{b1}, x_{w1}, t_1) \\
    (x_{b2}, x_{w2}, t_2) \\
    \vdots \\
    (x_{bn}, x_{wn}, t_n) \\
\end{bmatrix} := \text{modifiedNAGS}(G, n)$
\ForAll{$i \in 1 \dots n$} \Comment{can be run in parallel}
    \State $(\text{cost}_i, Q_i^\star) := \text{TrajOpt}
    (x_{bi}, x_{wi}, t_i, dt, T)$
\EndFor
\State $i^\star := \arg \min_i (\text{cost}_i)$
\State \Return $Q^\star_{i^\star}$
\EndFunction
\end{algorithmic}
\end{algorithm}

\subsection{Configuration Graph Generation}
\label{sec:graph}
The goal of this step is to transform the constrained high-dimensional continuous space of collision-free robot configurations into an unconstrained low-dimensional discrete graph for the subsequent NAGS algorithm.
The base heading and nonholonomic constraints are ignored at this stage.
The dimensionality reduction is achieved by a change of coordinates from $[x_b, x_w, x_e]^T \in \mathbb{R}^2 \times \mathbb{R}^3 \times \mathbb{R}^3$ to $[x_b, k, w]^T \in \mathbb{R}^2 \times [0, 1] \times \{0, 1\}$ by noticing that $x_e$ is fully defined by the path parameter $k$ and that given $x_b$ and $x_e$,
there only exists two feasible elbow positions: elbow up and elbow down, represented by $w=1$ and $w=0$ respectively, with $w \in \{0, 1\}$.
This is a direct results from Eq. (\ref{eq:kinematics}).
This parametrization reduces the configuration space dimensionality allowing for a simpler configuration graph and thus better runtime performance.
In the remainder of this section, we abuse notation and use $[x, y, k, w]^T$ and $[x_b, x_w, x_e]^T$ interchangeably with the understanding that the former can always be mapped to the latter via standard IK procedures \cite{siciliano2008springer}.

% Let $g: [x, y, t, w]^T \to [x_b, x_w, x_e]^T$ be the mapping function between the state representations.

% \begin{figure}[tbp]
%   \centering
%   \begin{subfigure}[b]{0.49\columnwidth}
%     \centering
%     \includegraphics[width=0.8\columnwidth]{figures/elbow_down.png}
%     \caption{Elbow down configuration}
%   \end{subfigure}%
%   ~
%   \begin{subfigure}[b]{0.49\columnwidth}
%     \centering
%     \includegraphics[width=0.8\columnwidth]{figures/elbow_up.png}
%     \caption{Elbow up configuration}
%   \end{subfigure}
%   \caption{Elbow up and elbow down configurations}
%   \label{fig:elbow_up_down}
% \end{figure}
The configuration graph (CG) is given by $G = (V, E)$ where $V$ is the set of vertices and $E$ is the set of undirected edges.
Beginning with $V$, vertices are defined via a discretization of the configuration space $(x, y, t, w) \in \mathbb{R}^2 \times [0, 1] \times \{0, 1\}$ by predefined discretization intervals $\Delta x, \Delta y, \Delta t$.
% This discretization determines the resolution of the graph and should be chosen based on the size of the smallest obstacle.
% The set of base positions $(x, y) \in \mathbb{R}^2$ which were originally unbounded, is also replaced by a bounded $(x, y) \in [-x_{\text{max}}, x_{\text{max}}] \times [-y_{\text{max}}, y_{\text{max}}]$ for some $x_{\text{max}}$ and $y_{\text{max}}$, defining the bounds of the base position.
We define the discretized bounded configuration space as
\begin{align*}
C &:= \{x_{\text{min}}, x_{\text{min}} + \Delta x, \dots, x_{\text{max}}\} \\
&\times
\{y_{\text{min}}, y_{\text{min}} + \Delta y, \dots, y_{\text{max}}\} \\
&\times \{t : 0, \Delta t, \dots, 1\} \times \{0, 1\}
\end{align*}
Define $C_{\text{free}} \subseteq C$ to be the configurations not in collision with obstacles.
Furthermore, given the upperarm link length $l_1$ and forearm link length $l_2$, the distance between the base and end effector cannot be greater than the full arm length ($l_1 + l_2$), thus we define
\begin{align*}
    C_\text{kinematic} = \{(x, y, t, w) : \lVert [x, y, 0]^T - x_e(t) \rVert _2 \leq l_1 + l_2\}
\end{align*}
The set of vertices is then given by $V = C_{\text{free}} \cap C_\text{kinematic}$.
% \begin{align*}
% V = C_{\text{free}} \cap C_\text{kinematic}
% \end{align*}

% Notice that for each $t$, the set of points $[x, y, 0]$ constitutes a \emph{disk} centered at $x_e(t)$ with radius $l_1 + l_2$.

% \begin{figure}[tbp]
% \smallskip

%   \begin{subfigure}[b]{0.5\linewidth}
%     \centering
%     \includegraphics[width=\textwidth]{figures/kings_graph_edges.png}
%     \caption{}
%     \label{fig:2d_king's_graph}
%   \end{subfigure}%
%   ~
%   % \centering
%   % \begin{subfigure}[b]{0.23\linewidth}
%   %   \centering
%   %   \includegraphics[width=\textwidth]{figures/3d_kings_graph.png}
%   %   \caption{}
%   % \end{subfigure}%
%   % ~
%   \centering
%   \begin{subfigure}[b]{0.3\linewidth}
%     \centering
%     \includegraphics[width=\textwidth]{figures/3d_kings_graph_vertex_all.png}
%     \caption{ }
%     \label{fig:3d_king's_graph_vertex_all}
%   \end{subfigure}
%   \caption{
%   Illustration of king's graphs. (a) Connectivity of a vertex in a 2D king's graph. (b) Connectivity of a vertex in a 3D king's graph.
%   }
%   \label{fig:3d_king's_graph}
%   % \vspace{-0.2cm}
% \end{figure}

% Edges between vertices represent possible transitions between the robot configurations which is determined by the state discretization, robot kinematics and presence of obstacles.

The vertices within an elbow configuration $w$ are connected by collision-free edges in a grid-like fashion with diagonals for coordinates $x, y, t$.
Transitions between elbow configurations $w$ can only occur at the joint singularity, when $\lVert [x, y, 0]^T - x_e(t) \rVert_2 = l_1 + l_2$.
A simplified example is illustrated in Fig. \ref{fig:1d_configuration_graph}.
The edge cost is defined as the Euclidean distance between vertices in the $(x, y, t)$ coordinates.

The construction of a configuration graph for general mobile manipulators follows similarly by solving for each joint's possible cartesian positions given $x_b$ and $x_e$, according to their kinematic constraints in Eq. (\ref{eq:kinematics}).

\subsection{Modified Neighborhood Augmented Graph Search}
\label{sec:NAGS}

The next step is to generate topologically distinct guesses.
Our approach is based on a modified version of the NAGS algorithm \cite[Algorithm 1]{sahin2023topogeometrically}.
The original algorithm along with our modifications colored in \color{blue}blue\color{black}, \color{orange} orange \color{black} and \color{magenta}magenta\color{black}, is presented in Algorithm \ref{alg:modified_NAGS2}.

\begin{algorithm}
\caption{Modified NAGS Algorithm}
\label{alg:modified_NAGS2}
\begin{algorithmic}[1]
\Require
\Statex $q_s \in V$: Start configuration
\Statex $q_g \in V$: Goal configuration
\Statex $\mathcal{N}_G$: Neighbor/successor function for graph $G$
\Statex $\mathcal{C}_G: V \times V \to \mathbb{R}^+$: Cost function
\Statex $n_\text{req}$: Required number of homotopically distinct paths
\Statex \textcolor{blue}{@computePS: $V \times (V_N, E_N)$: parent set computation}
\Ensure $G_N$: Graph with costs and parent set for every vertex
\Function{searchNAG}{$q_s$, $q_g$, $\mathcal{N}_G$, $\mathcal{C}_G$, $n_\text{req}$}
% \State $\begin{aligned}
%     &V_N := \{ v_s \}, &E_N := \emptyset \\
%     &v_s := (q_s, \{ q_s \}), &g(v_s) := 0 \\
%     &Q := \{ v_s \}, &v := v_s \\
%     &n := 0
% \end{aligned}$
\State $V_N := \{ v_s \}$, \quad $E_N := \emptyset$
\State $v_s := (q_s, \{ q_s \})$, \quad $g(v_s) := 0$
\State $Q := \{ v_s \}$, \quad $v := v_s$
\State $n := 0$
\While{$Q \neq \emptyset \land n < n_\text{req}$}
    \State $v := (q, U) = \arg \min_{v' \in Q} g(v')$
    \State $Q = Q - v$
    \color{orange}
    % \If{$\exists w \in G_N : v \equiv w$}
    %     \State \textbf{continue} \Comment{avoid visiting existing vertex}
    % \EndIf
    % Using \textcolor with \State sometimes causes weird bugs in indentation
    \State $V_N = V_N \cup \{ v \}$ \Comment{add vertex when visiting}
    \State $E_N = E_N \cup \{ (v, v.\text{came\_from})\}$
    \color{black}
    \color{blue}
    \State \st{$U' = \text{computePNS}(v, (V_N, E_N))$}
    \State $U' = \text{computePS}(v, (V_N, E_N))$
    \color{black}
    \color{magenta}
    \State \st{\textbf{for all $q' \in \mathcal{N}_G(v)$ do}}
    % \ForAll{$q' \in \mathcal{N}_G(v) : \exists (w \in \mathcal{N}_G(v)) \equiv (q', U')$}
    % \ForAll{$q' \in \mathcal{N}_G(v) : \substack{\exists w \in \mathcal{N}_G(v) \\ w \equiv (q', U')}$}
    \ForAll{$
      q' \in \mathcal{N}_G(v) :
      \left\{
        \begin{aligned}
          &\exists w \in \mathcal{N}_G(v) \\
          & w \equiv (q', U')
        \end{aligned}
      \right.
    $}
    \Statex \Comment{handle equivalent vertices first}
    \color{black}
        \State $v' := (q', U')$
        \State $g' = g(v) + C_G(q, q')$
        \State $w = v'$
        \State $E_N = E_N \cup \{ (v, w) \}$
        \If{$g' < g(w) \land w \in Q$}
            \State $g(w) = g'$
            \State $w.\text{came\_from} = v$
            \State $w.U = U'$
        \EndIf
    \EndFor
    \color{magenta}
    \ForAll{remaining $q' \in \mathcal{N}_G(v)$}
    \color{black}
        \State $v' := (q', U')$ \Comment{guaranteed new vertices}
        \color{orange}
        \State \st{$V_N = V_N \cup \{ v' \}$} \Comment{do not add vertex here}
        \State \st{$E_N = E_N \cup \{ (v, v') \}$}
        \color{black}
        \State $g(v') = g'$
        \State $v'.\text{came\_from} = v$
        \State $Q = Q \cup \{ v' \}$
        \If{$q' = q_g$}
            $n = n+1$
        \EndIf
    \EndFor
\EndWhile
\State \Return $G_N = (V_N, E_N)$
\EndFunction
\end{algorithmic}
\end{algorithm}

The main idea behind the original NAGS algorithm is to include approximations of the path tangents to vertices in Dijkstra's Algorithm \cite{dijkstra1959note}.
This is done by using a vertex's path neighborhood set (PNS)\cite[Algorithm 3]{sahin2023topogeometrically}, which is computed by running a reverse A* search \cite{hart1968formal} on the graph for a fixed search depth~$r$ from the current vertex back to the starting vertex.
Since homotopically distinct paths terminating at the same vertex should have distinct path tangents, vertices are considered distinct if their PNS do not intersect.
Note the distinction between CG vertex (vertices in the CG) and NAG vertex (vertices that are incrementally added to the NAG during the Dijkstra search).
Each NAG vertex corresponds to one CG vertex; however, multiple NAG vertices may correspond to the same CG vertex.
Two NAG vertices $v_1$ and $v_2$ are said to be \textit{coincident} if they correspond to the same CG vertex.
The equivalence ($\equiv$) of two coincident NAG vertices $v_1, v_2$ are thus defined as follows \cite[Definition 4]{sahin2023topogeometrically}
$$
v_1 \equiv v_2 \iff v_1\texttt{.cg} = v_2\texttt{.cg} \land v_1\texttt{.pns} \cap v_2 \texttt{.pns} \ne \emptyset
$$
where $v\texttt{.cg}$ and $v\texttt{.pns}$ retrieves NAG vertex $v$'s corresponding CG vertex and PNS, respectively.
We improve upon the original NAGS algorithm to address several shortcomings:

\subsubsection{Tiny obstacles}
\label{sec:tiny_obstacles}

\begin{figure}[tb]
  \begin{subfigure}[b]{0.15\columnwidth}
    \centering
    \includegraphics[width=0.9\columnwidth]{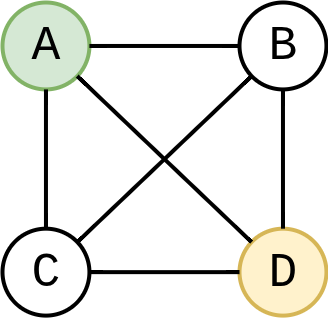}
    \caption{}
  \end{subfigure}%
  ~
  \begin{subfigure}[b]{0.15\columnwidth}
    \centering
    \includegraphics[width=0.9\columnwidth]{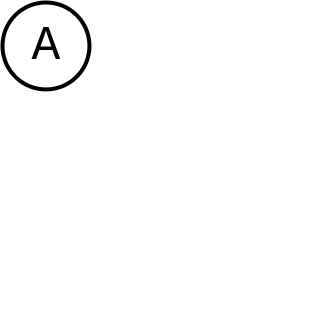}
    \caption{}
  \end{subfigure}%
  ~
  \begin{subfigure}[b]{0.15\columnwidth}
    \centering
    \includegraphics[width=0.9\columnwidth]{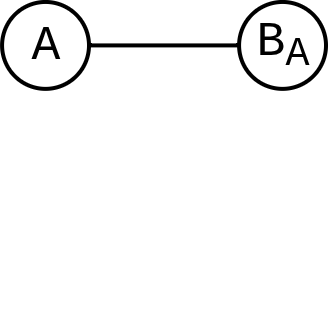}
    \caption{}
  \end{subfigure}%
  ~
  \begin{subfigure}[b]{0.15\columnwidth}
    \centering
    \includegraphics[width=0.9\columnwidth]{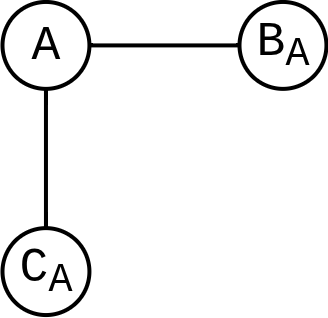}
    \caption{}
  \end{subfigure}%
  ~
  \begin{subfigure}[b]{0.15\columnwidth}
    \centering
    \includegraphics[width=0.9\columnwidth]{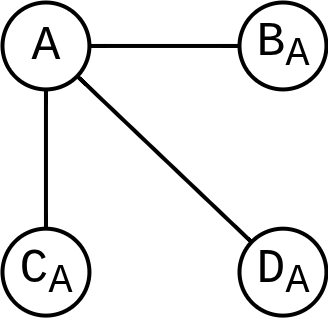}
    \caption{}
  \end{subfigure}%
  ~
  \begin{subfigure}[b]{0.15\columnwidth}
    \centering
    \includegraphics[width=0.9\columnwidth]{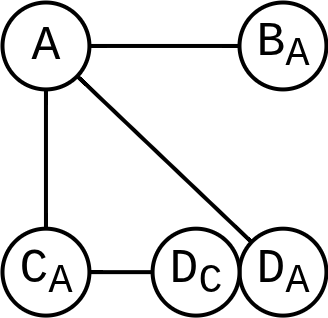}
    \caption{}
    \label{fig:modified_nags_motivation_r1_final}
  \end{subfigure}

  % \vspace{1mm}

  \begin{subfigure}[b]{0.15\columnwidth}
    \centering
    \includegraphics[width=0.9\columnwidth]{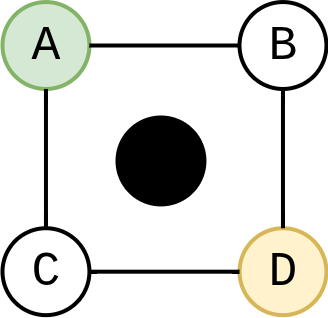}
    \caption{}
  \end{subfigure}%
  ~
  \begin{subfigure}[b]{0.15\columnwidth}
    \centering
    \includegraphics[width=0.9\columnwidth]{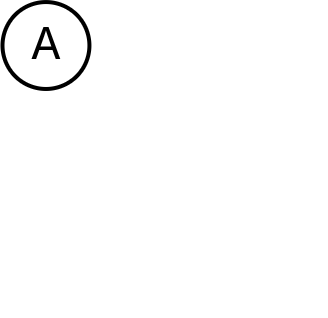}
    \caption{}
  \end{subfigure}%
  ~
  \begin{subfigure}[b]{0.15\columnwidth}
    \centering
    \includegraphics[width=0.9\columnwidth]{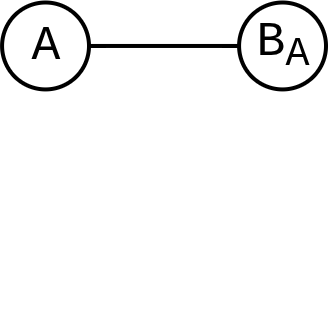}
    \caption{}
  \end{subfigure}%
  ~
  \begin{subfigure}[b]{0.15\columnwidth}
    \centering
    \includegraphics[width=0.9\columnwidth]{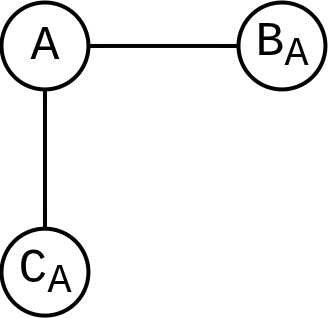}
    \caption{}
  \end{subfigure}%
  ~
  \begin{subfigure}[b]{0.15\columnwidth}
    \centering
    \includegraphics[width=0.9\columnwidth]{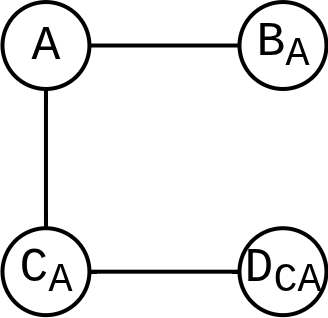}
    \caption{}
  \end{subfigure}%
  ~
  \begin{subfigure}[b]{0.15\columnwidth}
    \centering
    \includegraphics[width=0.9\columnwidth]{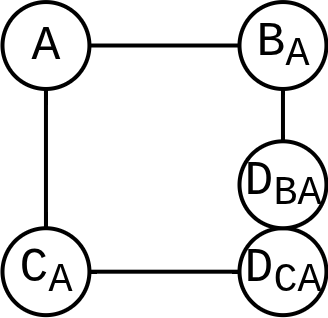}
    \caption{}
    \label{fig:modified_nags_motivation_r2_final}
  \end{subfigure}

  \caption{Top row: (a) shows a simple CG with the green start vertex and yellow goal vertex.  The edge weight corresponds to the length depicted.  Note that this CG only has one homotopically unique path from start to goal. (b)-(f) corresponds to the successive iterations as the NAG grows using $r = 1$.  The subscript indicates the PNS of the NAG vertex.  Notice in (f) that since NAG vertex $D_C$ and $D_A$ have disjoint PNS, NAGS incorrectly identifies them as homotopically distinct.
  Bottom row: (g) shows a simple CG with an obstacle in the middle.  Notice that there are two homotopically distinct paths from the start to goal.  (h)-(l) corresponds successive iterations with $r = 2$.  Notice in (l) that the NAG vertices $D_{BA}$ and $D_{CA}$ have overlapping PNS, thus NAGS incorrectly identifies them as homotopically equivalent.}
  \label{fig:modified_nags_motivation_r}
\end{figure}

The original NAGS algorithm suffers from the inability to distinguish homotopically distinct paths around tiny obstacles regardless of the value of $r$ chosen.
This is exemplified in Fig. \ref{fig:modified_nags_motivation_r}.
Notice that regardless of the value of $r$, the original NAGS algorithm is not able to correctly identify the homotopically distinct paths in the top and bottom cases simultaneously.
This stems from the fact that the continuous path tangent is poorly represented in a discrete graph structure.
As such, the overlap in the path tangent, approximated by the overlap of the PNS, is easily over- or underestimated, causing the incorrect identification of homotopically distinct paths.

\begin{figure}[tb]
  \begin{subfigure}[b]{0.24\columnwidth}
    \centering
    \includegraphics[width=\columnwidth]{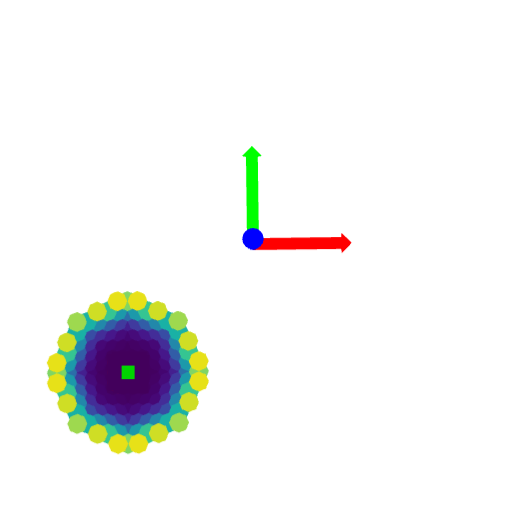}
    \caption{}
  \end{subfigure}%
  ~
  \begin{subfigure}[b]{0.24\columnwidth}
    \centering
    \includegraphics[width=\columnwidth]{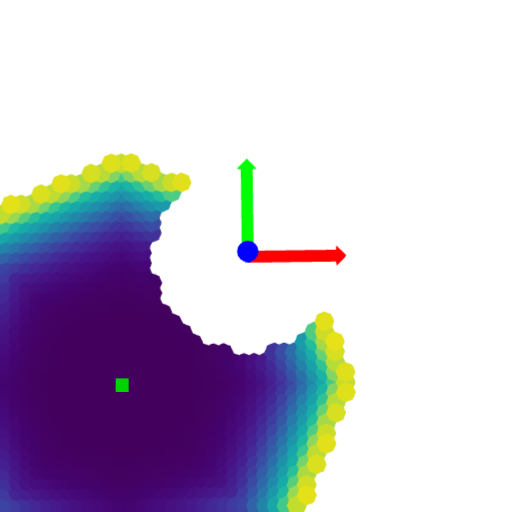}
    \caption{}
  \end{subfigure}%
  ~
  \begin{subfigure}[b]{0.24\columnwidth}
    \centering
    \includegraphics[width=\columnwidth]{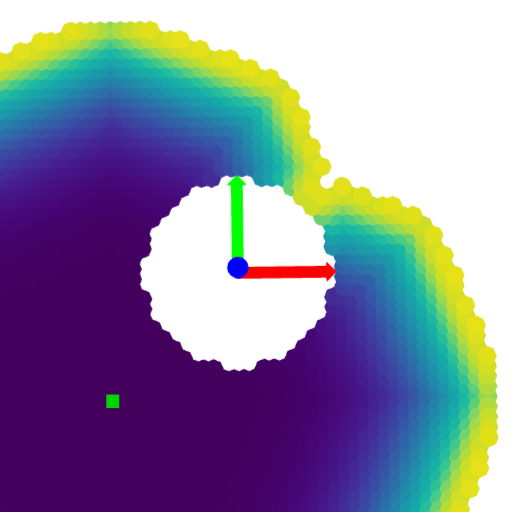}
    \caption{}
  \end{subfigure}%
  ~
  \begin{subfigure}[b]{0.24\columnwidth}
    \centering
    \includegraphics[width=\columnwidth]{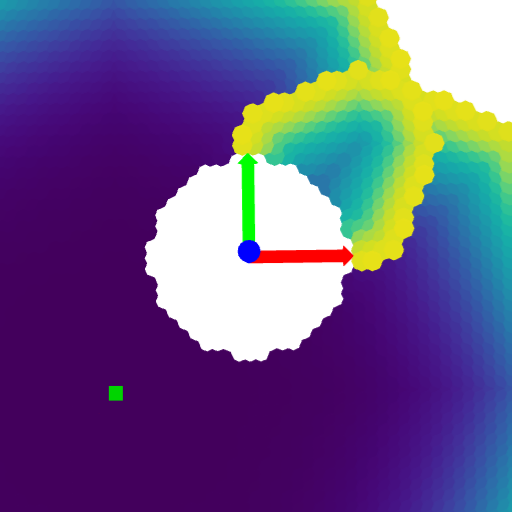}
    \caption{}
  \end{subfigure}%
  \caption{Successive iterations of the NAG.  Notice that obstacles cause the yellow wavefront to split and then merge.}
  \label{fig:nags_progression}
\end{figure}

\begin{figure}[tb]
  \begin{subfigure}[b]{0.24\columnwidth}
    \centering
    \includegraphics[width=0.6\columnwidth]{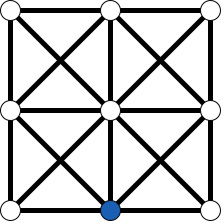}
    \caption{}
  \end{subfigure}%
  ~
  \begin{subfigure}[b]{0.24\columnwidth}
    \centering
    \includegraphics[width=0.6\columnwidth]{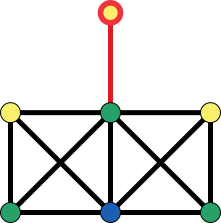}
    \caption{}
  \end{subfigure}%
  ~
  \begin{subfigure}[b]{0.24\columnwidth}
    \centering
    \includegraphics[width=0.6\columnwidth]{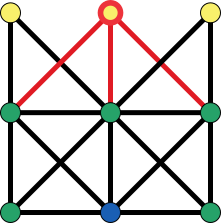}
    \caption{}
  \end{subfigure}%
  ~
  \begin{subfigure}[b]{0.24\columnwidth}
    \centering
    \includegraphics[width=0.6\columnwidth]{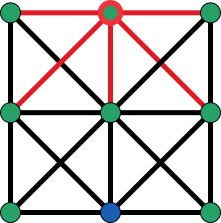}
    \caption{}
  \end{subfigure}%

    \begin{subfigure}[b]{0.24\columnwidth}
    \centering
    \includegraphics[width=0.6\columnwidth]{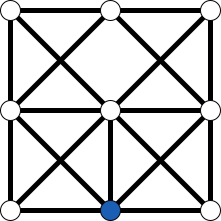}
    \caption{}
  \end{subfigure}%
  ~
  \begin{subfigure}[b]{0.24\columnwidth}
    \centering
    \includegraphics[width=0.6\columnwidth]{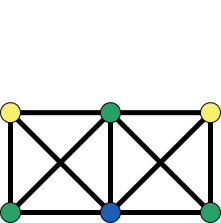}
    \caption{}
  \end{subfigure}%
  ~
  \begin{subfigure}[b]{0.24\columnwidth}
    \centering
    \includegraphics[width=0.6\columnwidth]{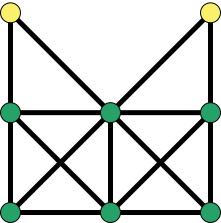}
    \caption{}
  \end{subfigure}%
  ~
  \begin{subfigure}[b]{0.24\columnwidth}
    \centering
    \includegraphics[width=0.6\columnwidth]{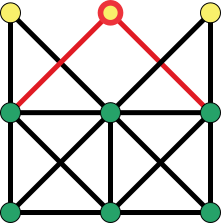}
    \caption{}
  \end{subfigure}%
  \caption{Top row: (a) shows a CG with the blue start vertex.
  (b)-(d) shows the successive iterations of the NAG.
  Yellow vertices are those in the open set (wavefront) while green vertices have already been visited.
  Consider the NAG vertex circled in red.
  The red edge connects the circled vertex to its parents.
  Bottom row: (e) shows a CG with an obstacle.
  (f)-(h) shows successive iterations of the NAG.
  }
  \label{fig:nags_ps_progression}
\end{figure}

Instead of a static approximation of the path tangent, we improve performance by using a dynamic local description of the wavefront of the open set instead (Fig. \ref{fig:nags_progression}).
The following observation can be made upon careful inspection of the wavefront as it visits a vertex.
Given the current state of the NAG $G_N = (V_N, E_N)$, define the parent set $\mathcal{P}(v)$ of NAG vertex $v$ as
\begin{align*}
\mathcal{P}(v) := \{v' : \exists (v, v') \in E_N \}
\end{align*}
We observe that in the absence of obstacles, the parent set~(PS) of a vertex starts on the shortest path and grows to adjacent vertices.
In the presence of obstacles, this rule is broken.
The PS no longer grows to adjacent vertices.
This is demonstrated in Fig. \ref{fig:nags_ps_progression}.
In effect, the PS acts as a local description of the wavefront.

This motivates the use of the PS to detect whether two paths are homotopically distinct or not.
% This is illustrated in Fig. \ref{fig:ps_progression}.
Two PS $\mathcal{P}_1, \mathcal{P}_2$ are said to be \textit{adjacent} if the following holds:
$$
\text{adj}_{E_N}(\mathcal{P}_1, \mathcal{P}_2) \iff \exists v_1 \in \mathcal{P}_1, v_2 \in \mathcal{P}_2 : (v_1, v_2) \in E_N
$$
We then redefine the equivalence relation ($\equiv$) as follows:
\begin{definition}[Equivalence between NAG vertices]
\label{def:equivalence}
For coincident NAG vertices $v_1$ and $v_2$ and NAG $G_N = (V_N, E_N)$,
$$
v_1 \equiv v_2 \iff
v_1\texttt{.cg} = v_2\texttt{.cg} \land \text{adj}_{E_N}(\mathcal{P}(v_1), \mathcal{P}(v_2))
$$
where $v\texttt{.cg}$ retrieves the corresponding CG vertex of $v$.
\end{definition}

To illustrate the effect of using PS, consider the example in Fig. \ref{fig:modified_nags_motivation_r1_final} again.
Using PS, $\mathcal{P}(D_C) = \{C_A\}, \mathcal{P}(D_A) = \{A\}$.
Since $C_A$ is adjacent $A$, $D_C \equiv D_A$ and they are considered homotopically identical.
For Fig. \ref{fig:modified_nags_motivation_r2_final}, $\mathcal{P}(D_{CA}) = \{C_A\}, \mathcal{P}(D_{BA}) = \{B_A\}$.
Since $C_A$ and $B_A$ are not adjacent, the two NAG vertices will be considered distinct, correctly identifying the two homotopically distinct paths.

\subsubsection{Non-uniform discretization}
\begin{figure}[tb]
  \begin{subfigure}[b]{0.24\columnwidth}
    \centering
    \includegraphics[width=\columnwidth]{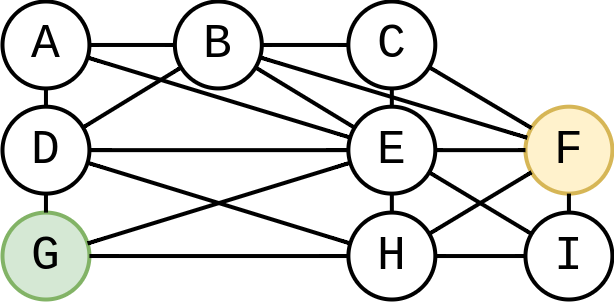}
    \caption{CG}
    \label{fig:bad_example1_cg}
  \end{subfigure}%
  \hfill  
  \begin{subfigure}[b]{0.24\columnwidth}
    \centering
    \includegraphics[width=0.9\columnwidth]{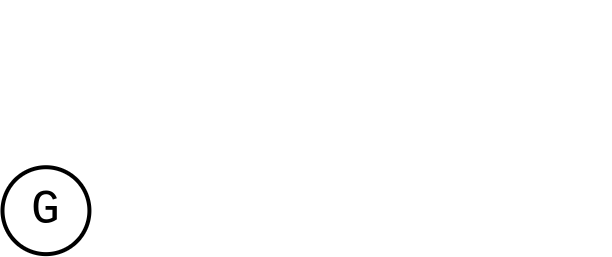}
    \caption{Start}
  \end{subfigure}%
  \hfill
  \begin{subfigure}[b]{0.24\columnwidth}
    \centering
    \includegraphics[width=\columnwidth]{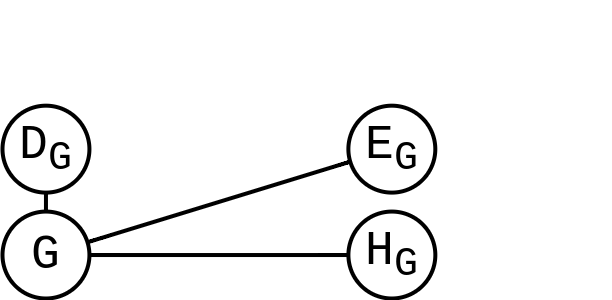}
    \caption{Visit G}
  \end{subfigure}%
  \hfill
  \begin{subfigure}[b]{0.24\columnwidth}
    \centering
    \includegraphics[width=\columnwidth]{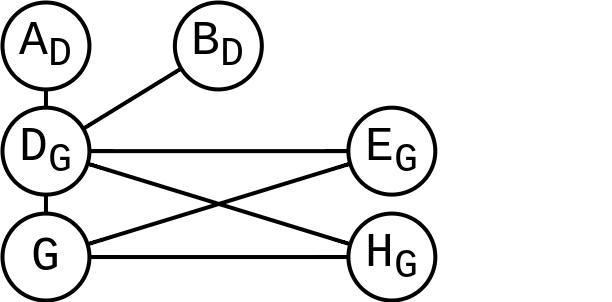}
    \caption{Visit D}
  \end{subfigure}

  \begin{subfigure}[b]{0.24\columnwidth}
    \centering
    \includegraphics[width=\columnwidth]{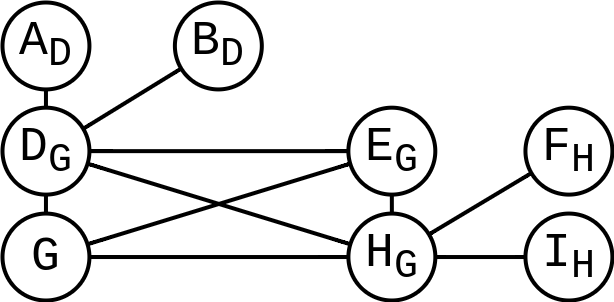}
    \caption{Visit H}
  \end{subfigure}%
  \hfill
  \begin{subfigure}[b]{0.24\columnwidth}
    \centering
    \includegraphics[width=\columnwidth]{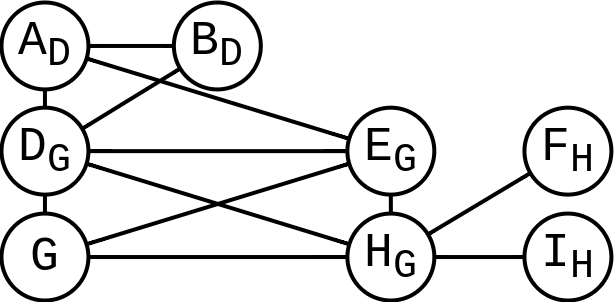}
    \caption{Visit A}
  \end{subfigure}%
  \hfill
  \begin{subfigure}[b]{0.24\columnwidth}
    \centering
    \includegraphics[width=\columnwidth]{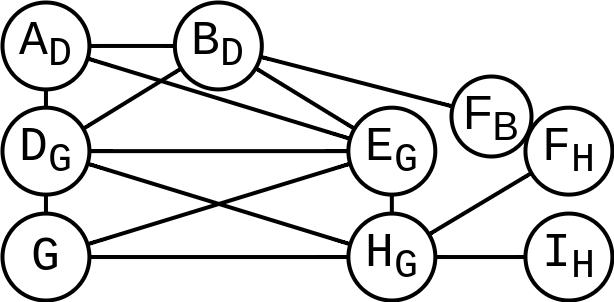}
    \caption{Visit B}
    \label{fig:bad_example1_6}
  \end{subfigure}%
  ~
  \caption{(a) Non-uniformly discretized CG with $GH<GD + DB<GE$.  Note that there is only 1 homotopically unique path between $G$ and $F$.
(b)-(g) Progression of the NAGS algorithm using PS.  Subscript indicates the parent of the vertex.  In (g), $\mathcal{P}(F_B) = \{B_D\}$, $\mathcal{P}(F_H) = \{H_G\}$.  Since $\mathcal{P}(F_B)$ is not adjacent to $\mathcal{P}(F_H)$, we have $F_B \not\equiv F_H$.
Hence the algorithm incorrectly determines that there are two homotopically distinct paths from $G$ to $F$.}
  \label{fig:bad_example1}
\end{figure}

In the original NAGS algorithm, $r$ must be fine tuned to account for potentially large differences in edge weights.
The PS modification mentioned in Section \ref{sec:tiny_obstacles} itself does not alleviate this issue.
One example is illustrated in Fig. \ref{fig:bad_example1}.
This is due to the fact that the original NAGS algorithm adds a vertex to the NAG based on the parent of that vertex.
We thus apply the changes in line 9-10 and line 25-26 of Algorithm \ref{alg:modified_NAGS2}.
These modifications ensure that vertices are added to the NAG in the order of the cost to the vertex itself, rather than the parent.
The effect of these modifications is that $F_E$ will be added to the NAG before $F_B$.
Since $F_E$ and $F_H$ are equivalent, the edge $(E_G, F_H)$ is inserted, causing the PS of $F_H$ to expand to $E_G$.
Then, $\mathcal{P}(F_H) = \{H_G, E_G\}$ will be adjacent to $\mathcal{P}(F_B) = \{B_D\}$ and thus $F_B \equiv F_H$.
% We also do an extra check for equivalence (line 11-12) to avoid reprocessing an existing vertex.

% \begin{figure}
%     \centering
%     \includegraphics[width=0.4\columnwidth]{figures/bad_example1_cg.png}
%     \caption{Non-uniformly discretized CG with $GH<GD + DB<GE$}
%     \label{fig:bad_example1_cg}
% \end{figure}

\subsubsection{Ambiguous visiting order}
% \begin{figure}
%     \centering
%     \includegraphics[width=0.24\columnwidth]{figures/bad_example2_cg.png}
%     \caption{}
%     \label{fig:bad_example2_cg}
% \end{figure}

\begin{figure}[tb]
  \begin{subfigure}[b]{0.18\columnwidth}
    \centering
    \includegraphics[width=0.95\columnwidth]{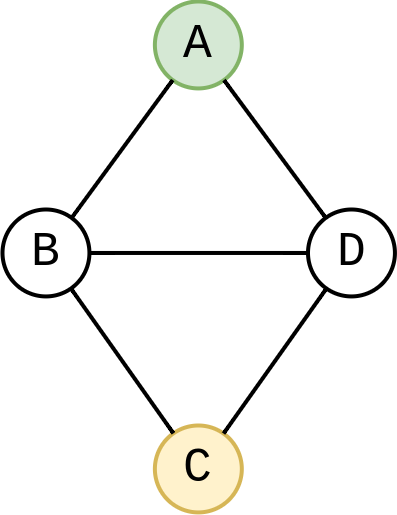}
    \caption{CG}
    \label{fig:bad_example2_cg}
  \end{subfigure}%
  \hfill
  \begin{subfigure}[b]{0.18\columnwidth}
    \centering
    \includegraphics[width=0.95\columnwidth]{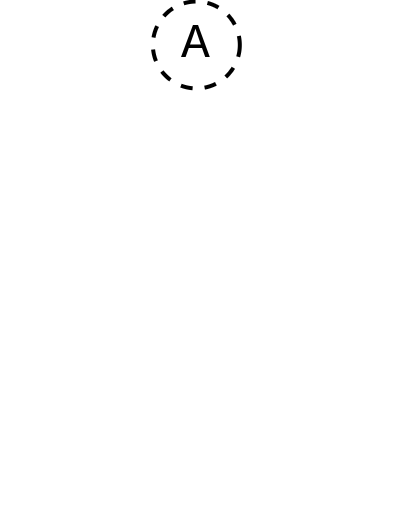}
    \caption{Start}
  \end{subfigure}%
  \hfill
  \begin{subfigure}[b]{0.18\columnwidth}
    \centering
    \includegraphics[width=0.95\columnwidth]{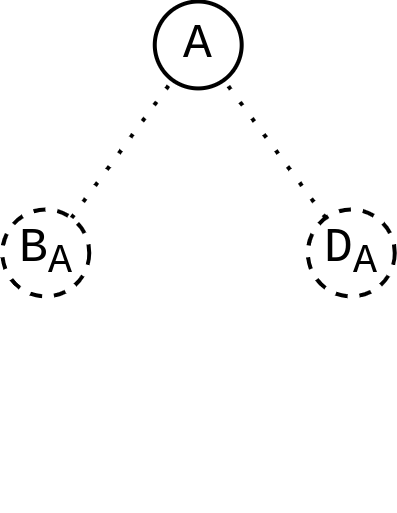}
    \caption{Visit A}
  \end{subfigure}%
  \hfill
  \begin{subfigure}[b]{0.18\columnwidth}
    \centering
    \includegraphics[width=0.95\columnwidth]{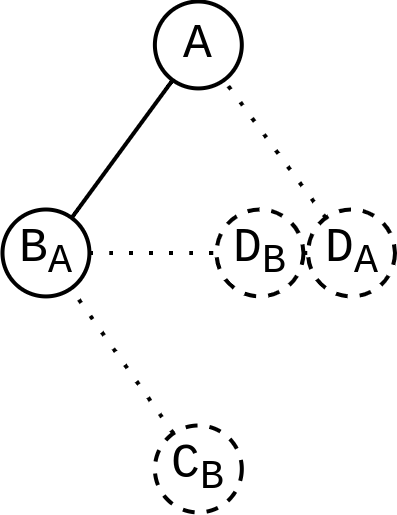}
    \caption{Visit B}
  \end{subfigure}%
  \hfill
  \begin{subfigure}[b]{0.18\columnwidth}
    \centering
    \includegraphics[width=0.95\columnwidth]{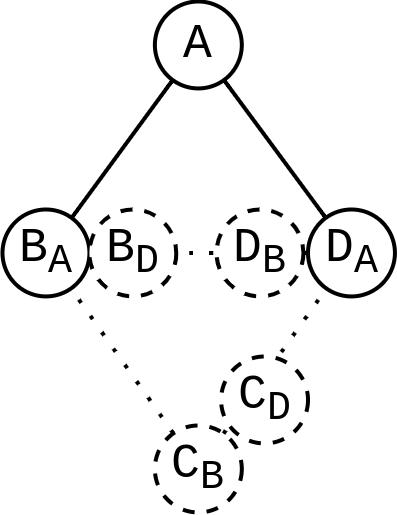}
    \caption{Visit D}
    \label{fig:bad_example2_4}
  \end{subfigure}
  \caption{(a) CG with uniform edge lengths.  Note that there is only 1 homotopically unique path between $A$ and $C$.
  (b)-(e) Progression of the NAGS algorithm using PS.  Subscript indicates the parent of the vertex. Dotted vertices represent vertices in the heap.
  Notice that the order for visiting $B_D$, $D_B$, $C_B$, $C_D$ is undefined as they all have the same path cost.
Furthermore, whether or not $C_B \equiv C_D$ depends on the order in which the four vertices are visited.
If $C_B$ and $C_D$ are visited before $B_D$ or $D_B$, then $C_B \not\equiv C_D$.}
  \label{fig:bad_example2}
\end{figure}

Similar to the above, the order in which vertices are visited impacts both PS and PNS calculation.
This is illustrated in Fig. \ref{fig:bad_example2}.
To tackle the nondeterministic visiting order of paths of the same length, we prioritize processing equivalent vertices first (line 13-14) before processing nonequivalent vertices (line 23).
This prevents equivalent vertices being incorrectly considered distinct due to visiting order.

% \begin{figure}[tbp]
%   \centering
%   \begin{subfigure}[b]{0.23\columnwidth}
%     \centering
%     \includegraphics[width=0.8\linewidth]{figures/nags_equiv.png}
%     \caption{}
%     \label{fig:nags_equiv}
%   \end{subfigure}%
%   ~
%   \centering
%   \begin{subfigure}[b]{0.23\columnwidth}
%     \centering
%     \includegraphics[width=0.8\linewidth]{figures/nags_distinct.png}
%     \caption{}
%     \label{fig:nags_distinct}
%   \end{subfigure}%
%   ~
%   \centering
%   \begin{subfigure}[b]{0.23\columnwidth}
%     \centering
%     \includegraphics[width=0.8\linewidth]{figures/pns_too_big.png}
%     \caption{}
%     \label{fig:pns_too_big}
%   \end{subfigure}%
%   ~
%   \centering
%   \begin{subfigure}[b]{0.23\columnwidth}
%     \centering
%     \includegraphics[width=0.8\linewidth]{figures/pns_too_small.png}
%     \caption{}
%     \label{fig:pns_too_small}
%   \end{subfigure}%
%   \caption{Two paths starting from the yellow vertex and ending at the green vertex, their PNSes marked black and white, along with a blue obstacle. (a) Homotopically equivalent paths have overlapping PNS. (b) Homotopically distinct paths have non-overlapping PNS. (c) Having $r$ too large causes false negatives. (d) Having $r$ too small causes false positives.}
%   \label{fig:pnses}
%   % \vspace{-0.35cm}
% \end{figure}

\subsubsection{Sufficient Condition Sketch}
We provide a more general sufficient condition for detecting homotopically distinct paths, only requiring obstacles that cause the removal of edges/vertices in the CG.

\begin{proposition}
Two locally shortest (geodesic) paths, $p_1$ and $p_2$, from the CG vertex $v_s$ to $v_g$ that encloses an obstacle will generate distinct NAG vertices.
\end{proposition}

\begin{sketchproof}
Note that each obstacle causes the removal of edges/vertices in the configuration graph, creating a chordless cycle $R_0$.
A chordless cycle is a cycle of length at least four in which no two vertices are joined by an edge outside of the cycle itself.
Define $R_k$ to be the set of vertices adjacent to but not contained in $R_{k-1}$.

Let $\{n_1, \dots, n_j\}$ be the NAG vertices of $p_1$ and $\{m_1, \dots, m_h\}$ be the NAG vertices of $p_2$.
Suppose path lengths $l(p_2) \ge l(p_1)$.
Note that $n_1$ and $m_1$ both correspond to the same CG vertex $v_s$ and $n_1 \equiv m_1$ while $n_j$ and $m_h$ both correspond to $v_g$, but it is yet to be determined whether they are equivalent or not.

The inductive proof proceeds as follows:
\begin{enumerate}
    \item Base condition: Consider $v_g \in R_0$.
    In order for $p_1$ and $p_2$ to enclose $R_0$ and be geodesics, we must have $n_{j-1}, m_{h-1} \in R_0$.
    Since $R_0$ is a chordless cycle, there is no edge between $n_{j-1}$ and $m_{h-1}$.
    $\mathcal{P}(n_j) = \{n_{j-1}\}$ is not adjacent $\mathcal{P}(m_h) = \{m_{h-1}\}$.
    Hence $n_j \not\equiv m_h$.
    \item Inductive step:  Assume the proposition is true $\forall v_g' \in R_{k-1}$.
    We wish to show it to be true for $v_g \in R_k$.
    Suppose the contrary, that for some $v_g \in R_k$, $n_j \equiv m_k$.
    This requires that $\mathcal{P}(m_{h}) = \{m_{h-1}\}$ be adjacent to $\mathcal{P}(n_{j}) = \{n_{j-1}\}$.
    We consider two cases for $m_{h-1}$
    \begin{enumerate}
        \item $m_{h-1} \in R_k \cup R_{k+1}$: This cannot be the case if $p_1$, $p_2$ enclose the obstacle, $l(p_2) \ge l(p_1)$ and they are locally shortest paths.
        This can be seen since there must exist some point $m_a \in p_2 \cap R_0$ and the subpath $(m_a, \dots, m_{h-1})$ is locally shortest.
        % \item $m_{k-1} \in R_k$: 
        \item $m_{h-1} \in R_{k-1}$: Then there exist some $v_g' = m_{h-1} \in R_{k-1}$ such that $p_1'$ and $p_2'$ are equivalent, contradicting the induction hypothesis.
    \end{enumerate}
    Hence the proposition must be true for $v_g \in R_k$.
\end{enumerate}
Thus we show the proposition to be true for pairs of locally shortest paths.
\end{sketchproof}
The generalization to all possible pairs of paths from $v_s$ to $v_g$ follows based on two facts: 1) If two NAG vertices $n_{j-1}$, $m_{h-1}$ are adjacent, then a path via $n_{j-1}$ to $m_{h-1}$ must have been deemed homotopically equivalent to a path via $m_{h-2}$ to $m_{h-1}$ in earlier iterations.
2) Dijkstra's algorithm always finds shorter paths first.
Thus, homotopic equivalence is determined by the locally shortest path.

\subsection{Trajectory Optimization}
\label{sec:traj_opt}
The goal of this step is to use the results from the previous section to refine the path, considering all constraints of the original planning problem.
In addition to including the base heading and the nonholonomic constraints, a finer time discretization is used to ensure that constraints are satisfied more precisely.

% We formulate the objective as finding the path that minimizes the motion of the base and elbow in 3D world coordinates/task space.
% Traditionally, it is more common to see trajectory optimization problems that minimize motion in joint space.
% Our departure is motivated by the following:
% Firstly, formulating in terms of base, elbow and end effector world coordinates avoids the need for the complex and highly nonconvex kinematics constraint for constraining end effector positions.
% Similarly, it also allows for more natural encoding of obstacle avoidance constraints since obstacles are also usually expressed in world coordinates.
% Furthermore, safety requirements often involve velocity constraints in world coordinates/task space rather than joint space.
% Using velocities in world coordinates opens the door for adding such safety critical constraints.
% Finally, we point out that joint motion is often proportional to motion of the elbow and base positions.  As such, minimizing the latter also approximately minimizes the former.
The trajectory optimization problem is given as
\allowdisplaybreaks
\begin{mini!}|s|
  {\substack{x_b[k], x_w[k], \theta[k]\\u_1[k], u_2[k], \Delta x_w[k]}} {
    \sum_{k=0}^T \lVert u_1[k] \rVert _2^2
    + \lVert u_2[k] \rVert _2^2
    + \lVert \Delta x_w[k] \rVert _2^2
  \label{opt:obj}}{}{}
  % \addConstraint{x_e[k]}{= x_e(kt) \forall k}
  \addConstraint{\lVert x_w[k] - x_b[k] \rVert _2 = l_1 \label{opt:const_l1}}
  \addConstraint{\lVert x_w[k] - x_e[k] \rVert _2 = l_2 \label{opt:const_l2}}
  \addConstraint{x_w[k] - x_b[k]= a(x_e[k] - x_b[k]) + \begin{bmatrix}0\\0\\b\end{bmatrix} \label{opt:elbow_kinematics}}
  \addConstraint{x_b[k+1]= x_b[k] + \begin{bmatrix}\cos \theta \\ \sin \theta \end{bmatrix} u_1[k]dt \label{opt:nonholonomic}}
  \addConstraint{\theta[k+1]= \theta[k] + u_2[k]dt \label{opt:angular_vel}}
  \addConstraint{x_w[k+1]= x_w[k] + \Delta x_w[k]dt \label{opt:elbow_vel}}
  \addConstraint{\text{obs}(x_b[k], x_w[k], x_e[k]) = \text{False} \label{opt:no_col}}
  % \addConstraint{(1 - \alpha_i) x_w[k] + \alpha_i x_b[k] \not \in \mathcal{C}_{\text{obs}}\label{opt:l1_nocol}}
  % \addConstraint{(1 - \alpha_i) x_w[k] + \alpha_i x_e[k] \not \in \mathcal{C}_{\text{obs}}\label{opt:l2_nocol}}
  %\addConstraint{|v[k]|}{\le v_{\text{max}} \label{opt:max_lin_vel}}
  %\addConstraint{|\omega[k]|}{\le \omega_{\text{max}} \label{opt:max_ang_vel}}
  %\addConstraint{|\Delta x_w[k]|}{\le \Delta x_{w\text{max}} \label{opt:max_elbow_vel}}
  %\addConstraint{x_b[0]} {= x_{b0} \label{opt:initial_xb}}
  %\addConstraint{x_w[0]} {= x_{w0} \label{opt:initial_xw}}
  %\addConstraint{\theta[0]} {= \theta_0 \label{opt:initial_theta}}
  %\addConstraint{x_b[T]} {= x_{bT} \label{opt:final_xb}}
  %\addConstraint{x_w[T]} {= x_{wT} \label{opt:final_xw}}
  %\addConstraint{\theta[T]} {= \theta_T \label{opt:final_theta}}
\end{mini!}
The objective (\ref{opt:obj}) is to minimize the discretized cost in Eq.~(\ref{eq:cost}).
This is subject to (\ref{opt:const_l1})-(\ref{opt:elbow_kinematics}) which enforce the kinematic constraints in Eq. (\ref{eq:kinematics}),
and (\ref{opt:nonholonomic})-(\ref{opt:elbow_vel}) which enforce the dynamic constraints in Eq. (\ref{eq:base_dynamics}) and (\ref{eq:arm_dynamics}).
Finally, (\ref{opt:no_col}) enforces collision avoidance at each timestep.

\subsection{Evaluate Local Optima}
% Given multiple distinct local optima from the previous step, the final step is then to comparing their respective final costs.
The final step is to compare and select the least cost path among the locally optimal paths from the previous step.
Formally, given the locally optimal trajectories $Q_i^\star$ and associated $\text{cost}_i$ for $i \in \{1, \dots, n\}$, index $i^\star$ of the trajectory with the least cost is given by
$$
i^\star := \arg \min_i(\text{cost}_i)
$$
The \emph{multi-locally optimal} path is then $Q^\star_{i^\star}$ which is the optimal path among the local optima $\{Q_1^\star, \dots, Q_n^\star\}$.
% This can be done in an offline or online fashion where one can wait until the specified number of local optima have been found and choose the one with minimal cost, or have a fixed time limit and only evaluate the local optima found so far.

\section{Results}
\label{sec:results}

% \begin{figure}[tbp]
%   \centering
%   \begin{subfigure}[b]{0.45\columnwidth}
%     \centering
%     \includegraphics[width=0.8\linewidth]{figures/planning_problem.png}
%     \caption{Planning Problem 1}
%     \label{fig:simple_planning_problem}
%   \end{subfigure}%
%   ~
%   \centering
%   \begin{subfigure}[b]{0.45\columnwidth}
%     \centering
%     \includegraphics[width=0.8\linewidth]{figures/table_planning_problem.png}
%     \caption{Planning Problem~2}
%     \label{fig:table_planning_problem}
%   \end{subfigure}%
%   \caption{The planning problems}
%   % \vspace{-0.2cm}
% \end{figure}

\begin{table*}[tb]
% \vspace{-0.4cm}
\begin{center}
\begin{tabular}{|l|l|l|l|l|l|l|}
\hline
\multicolumn{1}{|c|}{Method} & \multicolumn{3}{c|}{Planning Problem 1 (n=4)}   & \multicolumn{3}{c|}{Planning Problem~2 (n=3)} \\
\cline{2-7}
                         & Runtime(s) & NLP success rate      & Final cost   & Runtime(s) & NLP success rate & Final cost \\
\hline
\rowcolor{lightgray} CG+Modified NAGS (n)     & 1.02+0.49  & 1.0 / 1.0 / 1.0 / 1.0 & \textbf{2.33} / 6.02 / 6.09 / 6.05     & 0.90+2.35 & 1.0 / 1.0 / 1.0 & \textbf{2.32} / 2.98 / 2.99 \\
IMACS-KPIECE (1)         & 1.79    & 0.3                    & 14.37                                 & timeout & *               & *                  \\
IMACS-KPIECE (n)         & 29.9    & 0.6 / 0.2 / 0.2 / 0.1  & 10.57 / \textbf{5.75} / 25.64 / 10.03 & timeout & * / * / *       & * / * / *          \\
IMACS-RRTConnect (1)     & 37.3    & 0.3                    & 4.62                                  & timeout & *               & *                  \\
IMACS-RRTConnect (n)     & timeout & * / * / * / *          & * / * / * / *                         & timeout & * / * / *       & * / * / *          \\
Simple Interpolation (1) & 0.01    & 0.0                    & *                                     & 0.01 & 0               & *                  \\
\hline
\end{tabular}
\caption{Comparison between different initialization methods for the NLP. * indicates failure of current or prior steps.  (1) indicates singleshot mode. (n) indicates multishot mode, where $n$ is the number of $\mathcal{H}$-classes required.  The values separated by ``/'' are data corresponding to paths of each $\mathcal{H}$-class.  The bold text indicates the multi-locally optimal path}
\label{table:performance_comparisons}
\end{center}
% \vspace{-0.2cm}
%\vspace{-20pt}
\end{table*}

In this Section, we compare our method for generating topologically distinct initial guesses against sampling-based approaches and simple interpolation.
For all planning problems, we generate the CG by discretizing the end effector path at 0.05m intervals.
Base positions are discretized at a resolution of 0.1m.
Edges in the CG are further subsampled at 0.01m for collision checking.
% We note that while we generate the full CG in our experiments, using an implicit graph is also possible where the graph is generated on the fly as the NAGS algorithm is run.
The results of both modified NAGS and sampling-based methods are all used as initial guesses for the NLP specified in Section \ref{sec:traj_opt}.
The optimization problem is formulated in Drake \cite{drake} and solved with SNOPT \cite{gill2005snopt} using discretization $T = 200$ and timestep $dt = 0.2$.
% Obstacle avoidance is enforced with signed distances, similar to CHOMP \cite{zucker2013chomp} and TrajOpt \cite{schulman2014motion}.
All final costs represent optimized costs in Eq. (\ref{opt:obj}) after using the initial guess to solve the NLP.
Runtime only includes the time taken to generate the the initial guesses and does not include the time for solving the NLP .
The code for the implementation and comparison, as well as interactive recordings of the results can be found at
\url{https://github.com/rcywongaa/topologically_distinct_guesses}.
% \url{https://anonymous.4open.science/w/topologically_distinct_guesses-C0AA/}.
We investigate three planning scenarios.

\subsection{Two Sphere Obstacles with Straight Line Path Constraint}
\label{sec:planning_problem1}
Planning Problem~1 involves finding the optimal path for a simple two-link mobile manipulator in the presence of two spherical obstacles, subject to an end effector constraint in the form of a straight line.

Our algorithm is compared against two constrained sampling-based approaches: the IMACS-RRTConnect \cite{kuffner2000rrt} algorithm (emulates CBIRRT2 \cite{berenson2011constrained}, TB-RRT \cite{kim2016tangent}, AtlasRRT \cite{jaillet2012path}) and the IMACS-KPIECE \cite{csucan2009kinodynamic} algorithm which shows superior results in high dimensional constrained configuration space \cite{kingston2019exploring}.
The path tolerance was set to 0.05m.

% Both these methods use Implicit Manifold Configuration Space (IMACS) \cite{kingston2019exploring} to formulate the end effector constraints, allowing planners not originally developed for geometrically-constrained planning to be applied in this setting.
The sampling-based planners are compared in singleshot mode (1), where only one path is generated and evaluated, and multishot mode (n), where the planner keeps generating paths until at least one path belonging to each of the $n$ homotopic classes is generated.
Both modes are subject to a 5 minute timeout.
The sampling-based approaches were set up using OMPL \cite{sucan2012the-open-motion-planning-library} and MoveIt2 \cite{chitta2012moveit}\cite{coleman2014reducing}.
Additionally, a simple IK-based interpolation method is also compared.

\begin{figure}[tpb]
  \centering
  \begin{subfigure}[b]{0.23\columnwidth}
    \centering
    \includegraphics[width=\columnwidth]{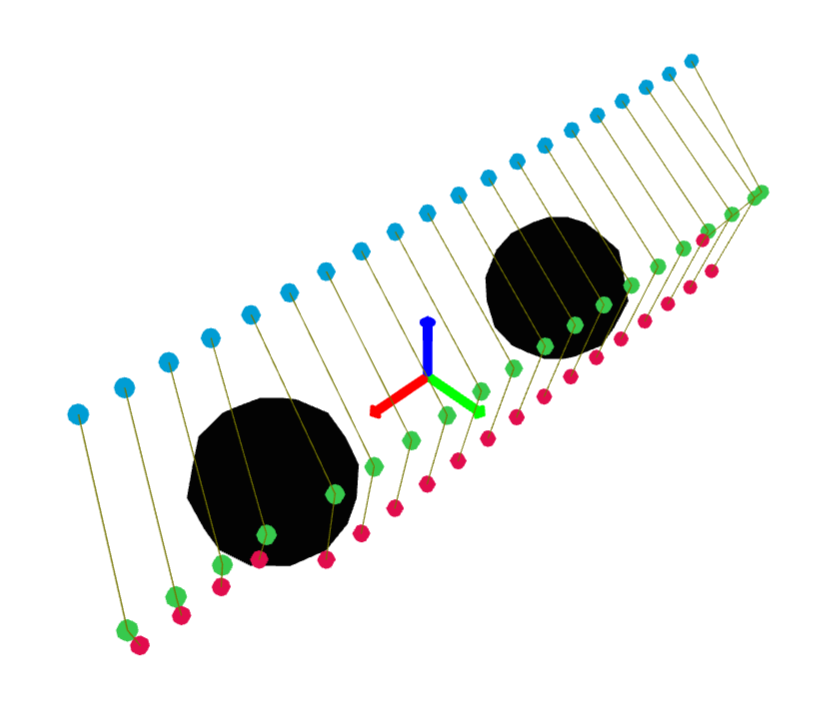}
    \caption{$\mathcal{H}$-class 1}
  \end{subfigure}%
  ~
  \begin{subfigure}[b]{0.23\columnwidth}
    \centering
    \includegraphics[width=\columnwidth]{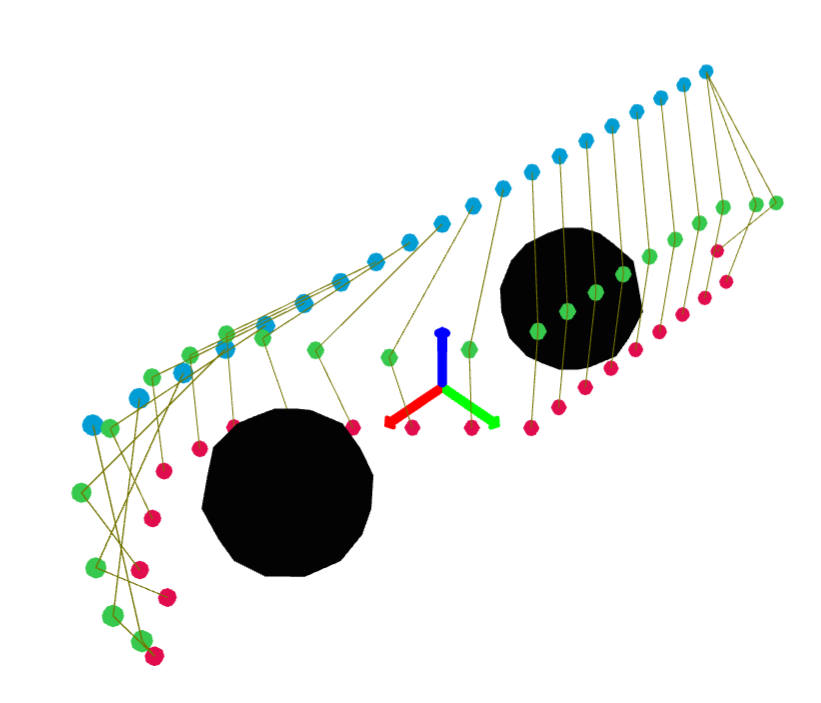}
    \caption{$\mathcal{H}$-class 2}
  \end{subfigure}%
  ~
  \begin{subfigure}[b]{0.23\columnwidth}
    \centering
    \includegraphics[width=\columnwidth]{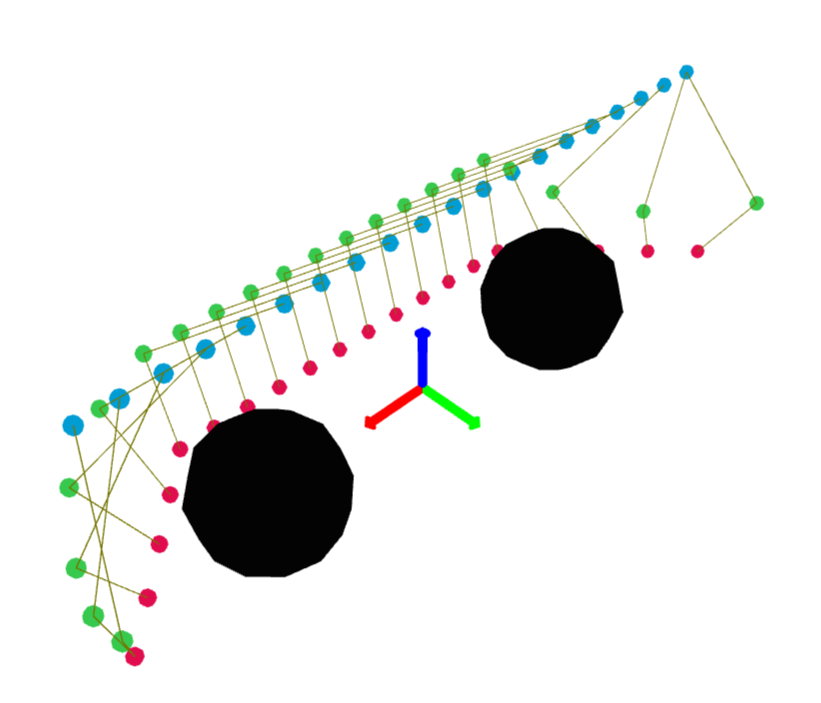}
    \caption{$\mathcal{H}$-class 3}
  \end{subfigure}%
  ~
  \begin{subfigure}[b]{0.23\columnwidth}
    \centering
    \includegraphics[width=\columnwidth]{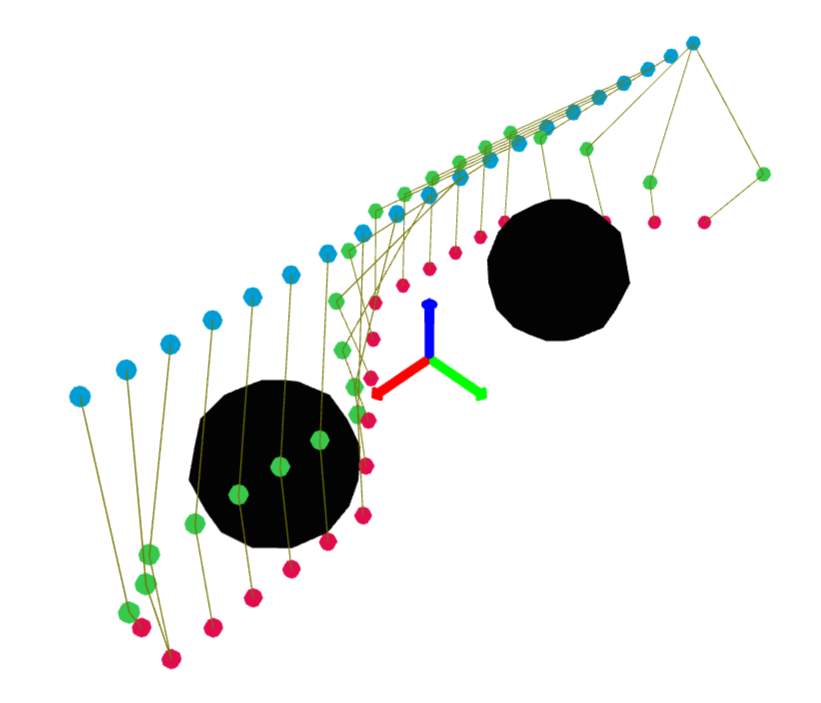}
    \caption{$\mathcal{H}$-class 4}
  \end{subfigure}%

  \caption{Results of NAGS belonging to different $\mathcal{H}$-classes for Planning Problem 1}
  \label{fig:NAG}
\end{figure}

\begin{figure}[tpb]
  \centering
  \begin{subfigure}[b]{0.23\columnwidth}
    \centering
    \includegraphics[width=\columnwidth]{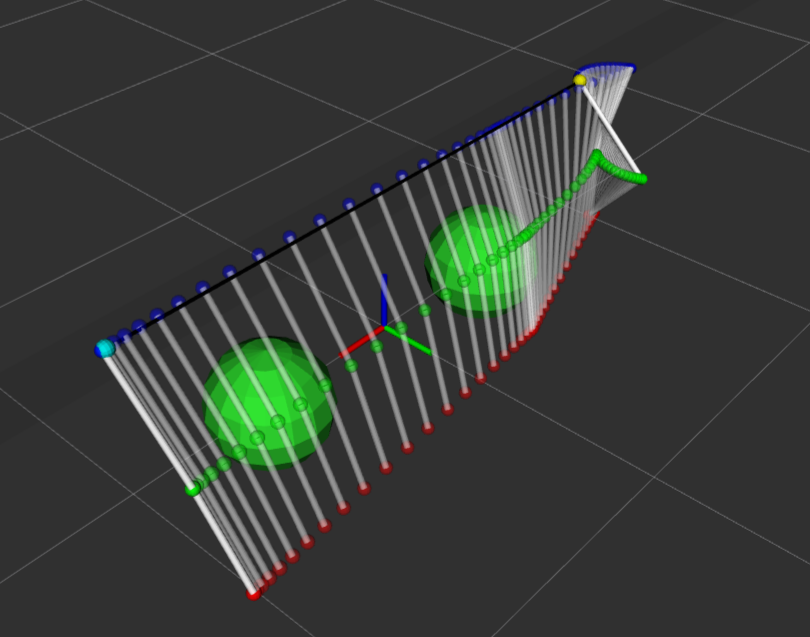}
    \caption{$\mathcal{H}$-class 1}
  \end{subfigure}%
  ~
  \begin{subfigure}[b]{0.23\columnwidth}
    \centering
    \includegraphics[width=\columnwidth]{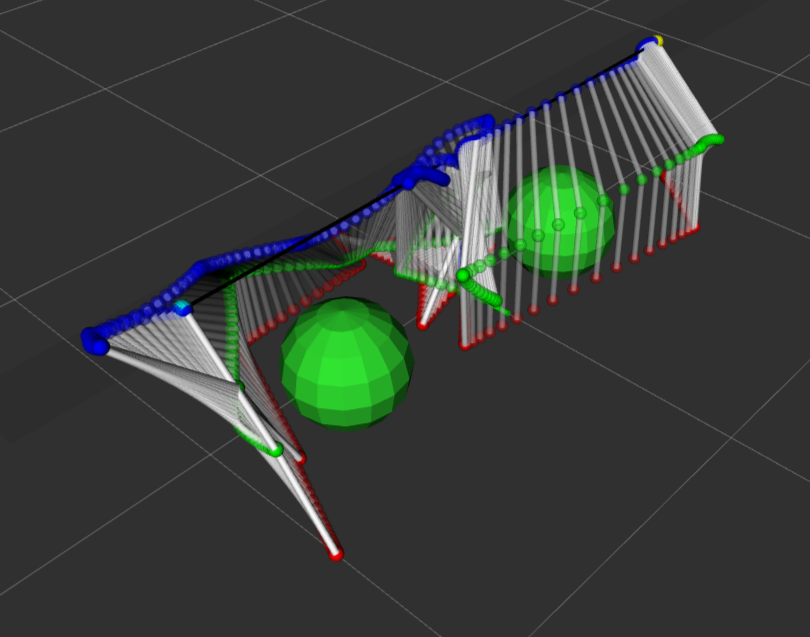}
    \caption{$\mathcal{H}$-class 2}
  \end{subfigure}%
  ~
  \begin{subfigure}[b]{0.23\columnwidth}
    \centering
    \includegraphics[width=\columnwidth]{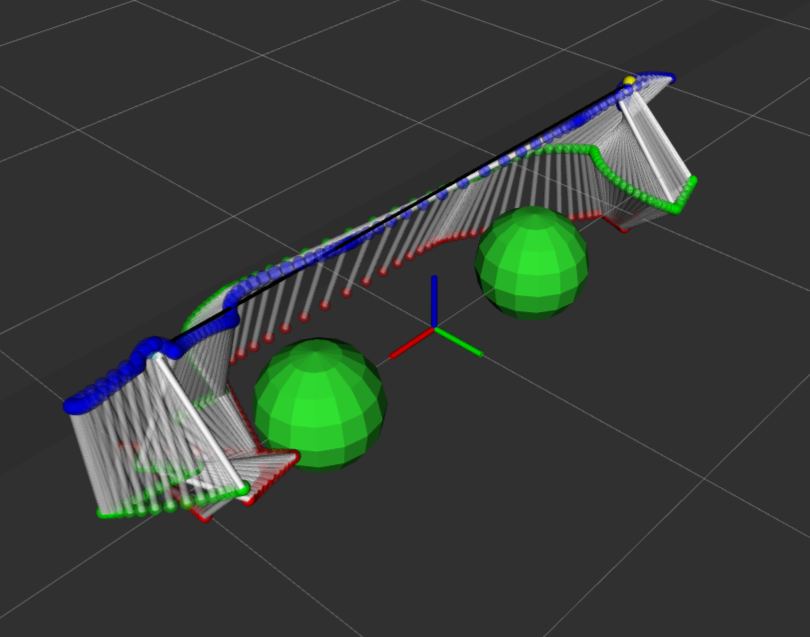}
    \caption{$\mathcal{H}$-class 3}
  \end{subfigure}%
  ~
  \begin{subfigure}[b]{0.23\columnwidth}
    \centering
    \includegraphics[width=\columnwidth]{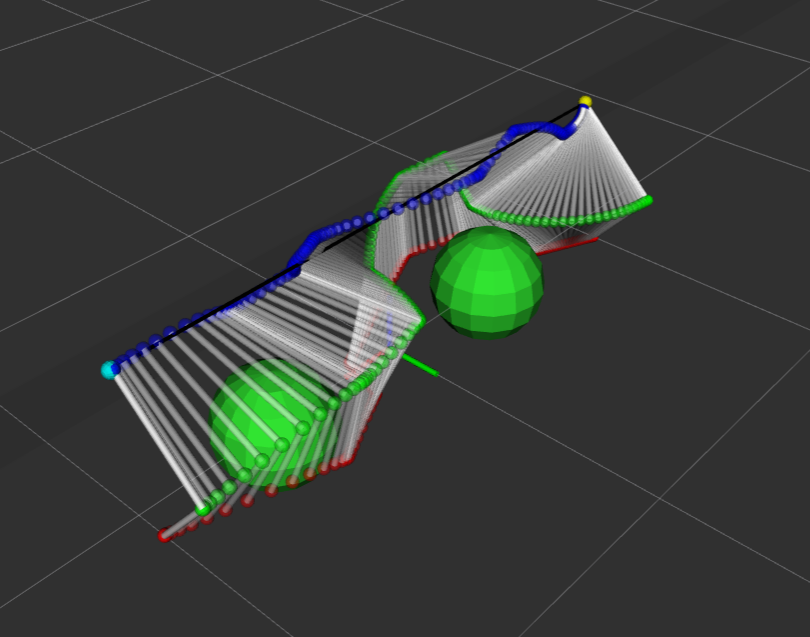}
    \caption{$\mathcal{H}$-class 4}
  \end{subfigure}%

  \caption{Results of IMACS-KPIECE belonging to different $\mathcal{H}$-classes for Planning Problem 1}
  \label{fig:OMPL}
    % \vspace{-0.6cm}
\end{figure}

The results are summarized in Table \ref{table:performance_comparisons}, averaged over 10 trials.
It can be seen that our algorithm can generate topologically distinct initial guesses more quickly as indicated by the lower runtime, as well as produce guesses of higher quality, as indicated by the higher NLP success rate and lower final cost.
Examples from modified NAGS and IMACS-KPIECE are shown in Fig. \ref{fig:NAG} and Fig.~\ref{fig:OMPL} respectively.

% Notice that in general, using topologically distinct paths as initial guesses allows for finding a multi-locally optimal path that with lower cost.
Note that the nonholonomic constraints of the mobile manipulator are only enforced in the NLP, further introducing more local optima to the optimization landscape.
This highlights the importance of providing high-quality guesses to the optimizer and explains the apparent differences in final cost of trajectories belonging to the same $\mathcal{H}$-classes.
Since the NAGS algorithm produces shortest paths within the $\mathcal{H}$-class, it naturally constitutes an initial guess that is closer to the global optimum, even with the added nonholonomic constraints.
Additionally worth pointing out, the single-shot performance of IMACS-KPIECE is comparable to that of our CG+Modified NAGS pipeline, while its multi-shot performance is significantly worse.
This indicates that the slowdown stems from the absence of homotopy awareness in IMACS-KPIECE.
Since the discovery of homotopically distinct paths is inherently probabilistic, this largely explains the increased planning time.

% The results also confirm the superiority of IMACS-KPIECE compared to IMACS-RRTConnect in high dimensional constraint motion planning as reported in \cite{kingston2019exploring}\cite{csucan2009kinodynamic}.

\subsection{Simulated Bar Table Cleaning}
Planning Problem~2 involves a more realistic table cleaning scenario in which a mobile manipulator in the form of a Kinova Gen3 robot arm attached to a Turtlebot 4 is tasked with cleaning a counter table with a sine wave motion while avoiding collisions.
The sampling-based planners are set up and evaluated in the same fashion as in Section \ref{sec:planning_problem1} and the results are summarized in Table \ref{table:performance_comparisons}, also averaged over 10 trials.
% The generated initial guess for Planning Problem~2 is shown in Fig. \ref{fig:NAG_table} and their corresponding optimized paths are shown in Fig. \ref{fig:intro}.
The final results are shown in Fig. \ref{fig:intro}.
Particularly, note that the highly nonconvex table and chair means that conventional approaches of projecting obstacles to the ground plane and splitting base and arm motion planning would yield poor results since naive projection would either severely overestimate or underestimate the size of the obstacles.
Indeed, if we project the table and chair to the ground plane and consider the mobile base inflation radius, the path where the base moves between the table and the chair would not be feasible.
This planning problem also highlights the limitation of sampling based planners which struggle with narrow passages generated by the large obstacle and end effector constraints \cite{kingston2019exploring}.

\begin{figure}[tpb]
  \centering
  \begin{subfigure}[b]{0.49\columnwidth}
    \centering
    \includegraphics[width=\columnwidth]{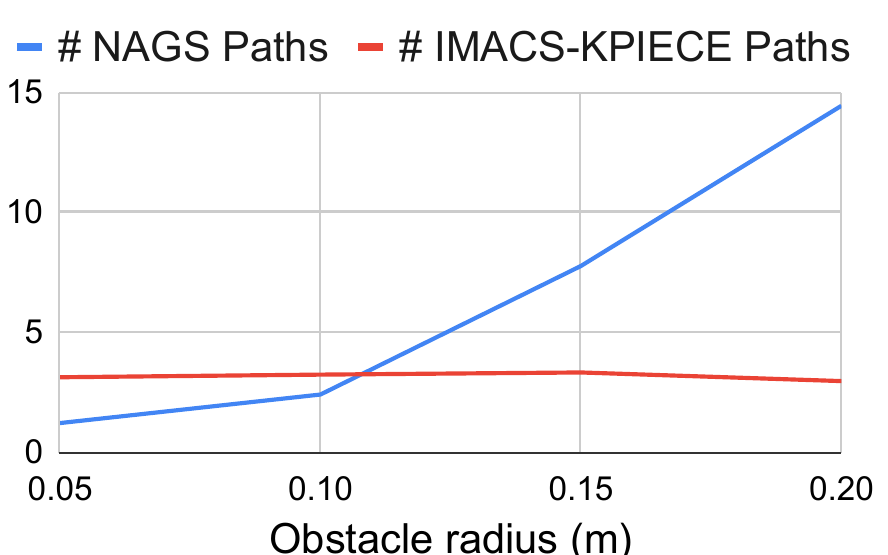}
    \caption{}
  \end{subfigure}%
  ~
  \begin{subfigure}[b]{0.49\columnwidth}
    \centering
    \includegraphics[width=\columnwidth]{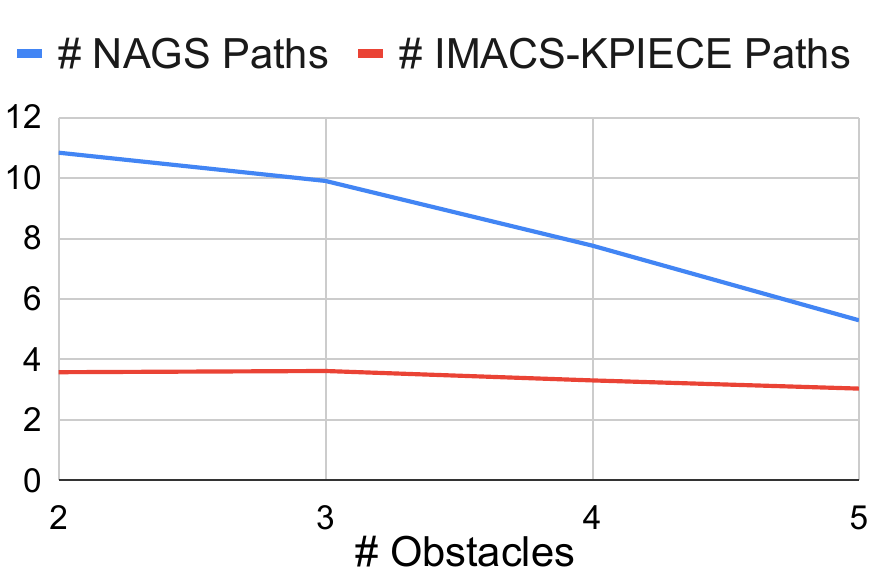}
    \caption{}
  \end{subfigure}%
  \caption{Effect of varying number of obstacles and obstacle radius on number of paths found in 10s.  Note that paths found by IMACS-KPIECE may not be homotopically distinct. (a) uses 4 obstacles. (b) uses obstacles of radii 0.15m.}
  \label{fig:randomized_tests}
\end{figure}

\subsection{Randomized Tests}
This test studies the effect of the number and size of the obstacles on the performance of our modified NAGS algorithm.
The setup is similar to Planning Problem~1 with a varying number and radii of spherical obstacles.
In this experiment, we focus on comparing against IMACS-KPIECE since it has been shown to be superior in performance in Planning Problem~1.
% to IMACS-RRTConnect in Planning Problem~1 and in \cite{kingston2019exploring,csucan2009kinodynamic}.
Since it is difficult to predetermine the number of $\mathcal{H}$-classes in a randomized setting,
we instead impose a 10s time limit for both modified NAGS and IMACS-KPIECE to generate as many paths as possible (homotopically distinct or not).
The experiment is conducted with randomized obstacle positions averaged over 50 trials per setting.
The results are shown in Fig. \ref{fig:randomized_tests}.
It can be seen that modified NAGS outperforms IMACS-KPIECE in situations with a small number of large obstacles.
Large obstacles are favorable since they lead to fewer vertices in the CG, and thus to a faster search.
For a high number of obstacles, the number of homotopically distinct paths of a certain length grows rapidly.
This causes the open set to grow quickly, which impacts the performance.
Furthermore, since each obstacle creates infinitely many homotopically distinct paths corresponding to increasing winding numbers, there is no upper bound to the size of the open set.

Using 4 obstacles of radius of 0.15, we repeat the experiment with the trajectory optimization step included.
We additionally randomize the initial and goal headings.
% Additionally, comparison with basic Dijkstra's Algorithm \cite{dijkstra1959note} on the same CG is performed.
The respective multi-locally optimal paths are then compared.
Across 50 trials, modified NAGS and IMACS-KPIECE produced at least one admissible initial guesses for 88\% and 60\% of the trials, respectively, where an initial guess is admissible if it allows the NLP to be solved.
On average, the multi-locally optimal paths generated by the NAGS algorithm were 30.6\% shorter than those generated by IMACS-KPIECE.
% Finally, in 34\% of the scenarios, NAGS provided initial guesses that resulted in a lower optimal path than that provided by basic Dijkstra.
% This shows that NAGS successfully found paths belonging to $\mathcal{H}$-class other than 
Furthermore, 34\% of the multi-locally optimal paths from modified NAGS were \emph{not} from the first (shortest) path returned.
Since the first path returned by modified NAGS is equivalent to the result of running standard Dijkstra's Algorithm \cite{dijkstra1959note}, this also shows that modified NAGS produced better initial guesses than standard Dijkstra in those cases.

\section{Conclusion \& Discussions}
\label{sec:conclusion}
% This paper presents a pipeline for nonholonomic mobile manipulator path planning in the presence of end effector constraints that achieve multi-local optimality.
% % This is achieved by discretizing a reduced state space into a configuration graph,  generating homotopically distinct paths via a modified NAGS algorithm,
% % using these distinct paths as initial guesses for the subsequent NLP program, and finally choosing the best among the returned distinct local optima.
% This is achieved by generating and optimizing homotopically distinct guesses and finally choosing the best among the local optima.
% Experiments showed that our pipeline was able to generate \emph{multi-locally optimal} solutions for a fairly complex table-cleaning scenario.
% Furthermore, the experiments showed our scheme for generating homotopically distinct paths outperformed existing constrained motion planning approaches.

This paper presents a pipeline for mobile
manipulator path planning under end effector path constraints that achieve multi-local optimality.
Several modifications were proposed to the core NAGS algorithm enabling it to reliably distinguish homotopically distinct paths.
Our algorithm performs particularly well in scenarios where kinematic structure and constraints reduce the dimensionality of the problem, and where a small number of large obstacles lead to a more compact configuration graph.
In such cases, the algorithm’s ability to handle large obstacles becomes especially beneficial, as these environments often introduce challenging local optima that our approach is well-equipped to address.
% As such, our pipeline generally excels in scenarios where kinematics and constraints can reduce the dimensionality of the problem and where a small number of large obstacles lead to a smaller configuration graph.
% Since NAGS will never find all homotopically distinct paths (there are infinitely many), future work may investigate methods for determining the correct number of homotopically distinct paths to find.
This can be seen as a complement to sampling-based approaches, which generally work well in the absence of constraints and with smaller, more numerous obstacles.
Future work may investigate ways to alleviate the curse of dimensionality when applying our algorithm to mobile manipulators with many DoFs.

%\clearpage
\bibliographystyle{IEEEtran}
\bibliography{IEEEabrv,references}

\end{document}